%% file: main.tex
\pdfoutput=1
\PassOptionsToPackage{T2A}{fontenc}
\documentclass[multiauthors,onecolumn]{cohere}

\usepackage{graphicx}
\usepackage{booktabs}
\usepackage{subcaption}
\usepackage{longtable}
\usepackage{pgfplots}
\usepackage{pgfplotstable}
\usepackage{tabularx}
\usepackage{xspace}
\usepackage{graphicx}
\usepackage{multirow}
\usepackage[normalem]{ulem}
\usepackage{pdfpages}
\usepackage{makecell}
\usepackage{fontawesome}
\usepackage{listings}
\usepackage{float}
\usepackage{xcolor}
\usepackage{rotating}
\usepackage{arydshln}
\usepackage{placeins}
\usepackage{amsmath}
\usepackage{footmisc}
\usepackage{tikz}

 \definecolor{ayadsymbol}{RGB}{76, 110, 230}
 \definecolor{secondsymbol}{RGB}{140, 165, 240}
 
 \DeclareSymbolFont{extraup}{U}{zavm}{m}{n}
 \DeclareMathSymbol{\vardiamond}{\mathalpha}{extraup}{87}
 \renewcommand{\fasymbol}{\textcolor{secondsymbol}{$\vardiamond$}}%

\newcommand{\benchmark}{\textsc{Kaleidoscope}\xspace}

\title{Kaleidoscope: In-language Exams for Massively Multilingual Vision Evaluation}

\author{name={Israfel Salazar}\faa,affiliation={2}}
\author{name={Manuel Fernández Burda}\faa,affiliation={3}}
\author{name={Shayekh Bin Islam}\faa,affiliation={4}}
\author{name={Arshia Soltani Moakhar}\faa,affiliation={4}}
\author{name={Shivalika Singh}\faa,affiliation={1}}
\author{name={Fabian Farestam}\faa,affiliation={17}}
\author{name={Angelika Romanou}\faa,affiliation={}}
\author{name={Danylo Boiko},affiliation={4,9}}
\author{name={Dipika Khullar},affiliation={4}}
\author{name={Mike Zhang},affiliation={10}}
\author{name={Dominik Krzemiński},affiliation={4}}
\author{name={Jekaterina Novikova},affiliation={4}}
\author{name={Luísa Shimabucoro},affiliation={18}}
\author{name={Joseph Marvin Imperial},affiliation={19,26}}
\author{name={Rishabh Maheshwary},affiliation={4}}
\author{name={Sharad Duwal},affiliation={4}}
\author{name={Alfonso Amayuelas},affiliation={5}}
\author{name={Swati Rajwal},affiliation={6}}
\author{name={Jebish Purbey},affiliation={4,7}}
\author{name={Ahmed Ruby},affiliation={25}}
\author{name={Nicholas Popovič},affiliation={11,12}}
\author{name={Marek Suppa},affiliation={22,23}}
\author{name={Azmine Toushik Wasi},affiliation={4}}
\author{name={Ram Mohan Rao Kadiyala},affiliation={7,8}}
\author{name={Olga Tsymboi},affiliation={13, 28}}
\author{name={Maksim Kostritsya},affiliation={15,16}}
\author{name={Bardia Soltani Moakhar},affiliation={4}}
\author{name={Gabriel da Costa Merlin},affiliation={18}}
\author{name={Otávio Ferracioli Coletti},affiliation={18}}
\author{name={Maral Jabbari Shiviari},affiliation={4}}
\author{name={MohammadAmin farahani fard},affiliation={4}}
\author{name={Silvia Fernandez},affiliation={4}}
\author{name={María Grandury},affiliation={21}}
\author{name={Dmitry Abulkhanov},affiliation={4}}
\author{name={Drishti Sharma},affiliation={4,7}}
\author{name={Andre Guarnier De Mitri},affiliation={18}}
\author{name={Leticia Bossatto Marchezi},affiliation={20}}
\author{name={Setayesh Heydari},affiliation={4}}
\author{name={Johan Obando-Ceron},affiliation={4,24}}
\author{name={Nazar Kohut},affiliation={14}}
\author{name={Beyza Ermis},affiliation={1}}
\author{name={Desmond Elliott}\textsuperscript{\fasymbol},affiliation={2,27}}
\author{name={Enzo Ferrante}\textsuperscript{\fasymbol},affiliation={3}}
\author{name={Sara Hooker}\textsuperscript{\fasymbol},affiliation={1}}
\author{name={Marzieh Fadaee}\textsuperscript{\fasymbol},affiliation={1}}

\affiliations{
    \item[1] Cohere For AI
    \item[2] Department of Computer Science, University of Copenhagen
    \item[3] Institute of Computer Sciences, CONICET \& Universidad de Buenos Aires
    \item[4] Cohere For AI Community
    \item[5] University of California, Santa Barbara
    \item[6] Emory University
    \item[7] M2ai.in
    \item[8] Traversaal.ai
    \item[9] Taras Shevchenko National University of Kyiv
    \item[10] Aalborg University
    \item[11] Karlsruhe Institute of Technology, Germany
    \item[12] ScaDS.AI, TU Dresden, Germany
    \item[13] T-Tech
    \item[14] Lviv Polytechnic National University
    \item[15] HSE University (Higher School of Economics)
    \item[16] RAFT
    \item[17] ETH Zürich
    \item[18] University of São Paulo
    \item[19] National University Philippines
    \item[20] Federal University of São Carlos
    \item[21] SomosNLP
    \item[22] Cisco
    \item[23] Comenius University in Bratislava
    \item[24] Mila, University of Montreal
    \item[25] Uppsala University
    \item[26] University of Bath
    \item[27] Pioneer Center for AI
    \item[28] Moscow Institute of Physics and Technology
}
\corresponding[*]{Israfel Salazar \texttt{<israfel.salazar@di.ku.dk>}, Manuel Fernández Burda \texttt{<mburda@dc.uba.ar>}, Marzieh Fadaee \texttt{<marzieh@cohere.com>}}

\newcommand{\tabitem}{~~\llap{\textbullet}~~}

\newcommand{\earthmoji}{\faGlobe}
\newcommand{\diversitymoji}{\faUsers}
\newcommand{\mmemoji}{\faPictureO}

\pgfplotsset{compat=1.18}

\begin{document}

\begin{abstract}

The evaluation of vision-language models (VLMs) has mainly relied on English-language benchmarks, leaving significant gaps in both multilingual and multicultural coverage. While multilingual benchmarks have expanded, both in size and languages, many rely on translations of English datasets, failing to capture cultural nuances. In this work, we propose \benchmark, as the most comprehensive exam benchmark to date for the multilingual evaluation of vision-language models. \benchmark is a large-scale, in-language multimodal benchmark designed to evaluate VLMs across diverse languages and visual inputs. \benchmark covers 18 languages and 14 different subjects, amounting to a total of 20,911 multiple-choice questions. Built through an open science collaboration with a diverse group of researchers worldwide, \benchmark ensures linguistic and cultural authenticity. We evaluate top-performing multilingual vision-language models and find that they perform poorly on low-resource languages and in complex multimodal scenarios. Our results highlight the need for progress on culturally inclusive multimodal evaluation frameworks. 

\small
\textbf{website:} \url{http://cohere.com/research/kaleidoscope}\\
\textbf{dataset:} \url{https://hf.co/datasets/CohereForAI/kaleidoscope}
\normalsize

\end{abstract}

\begin{figure}[thb!]
    \centering
    \includegraphics[width=1\textwidth]{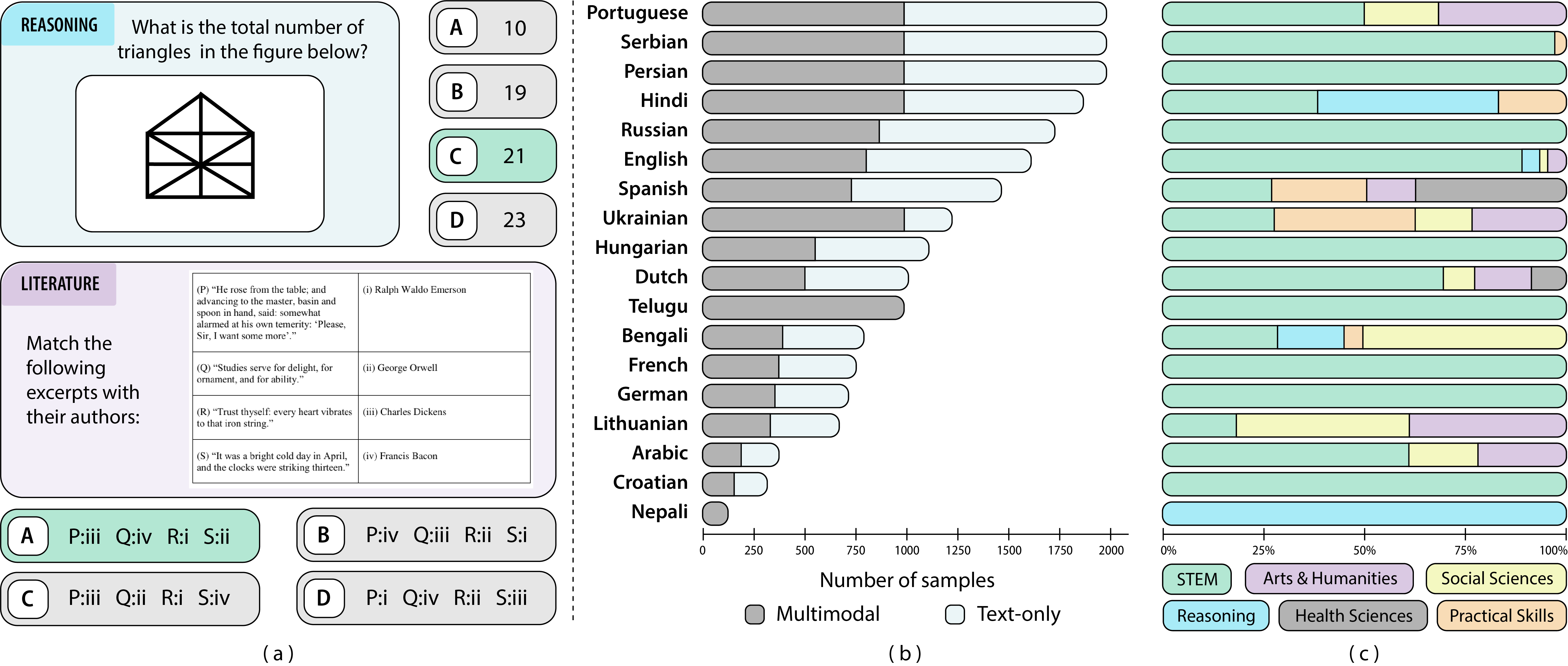}
    \caption{\textbf{Overview of the \textnormal{\benchmark} Benchmark.} (a) Multilingual-Multimodal MCQ Samples (b) Language and Multimodal Samples Distribution. (c) Exam Category Breakdown.}
    \label{fig:overview-figure}
\end{figure}

\section{Introduction}
\label{sec:introduction}
 
Evaluations are the backbone of measuring progress in machine learning, yet many benchmarks \---\ especially for language models \---\ continue to mirror an English and Western-centric worldview~\citep{joshi-etal-2020-state,fan2020englishcentricmultilingualmachinetranslation,dodge-etal-2021-documenting,liu-etal-2021-visually,chung2022scalinginstructionfinetunedlanguagemodels,gehrmann-etal-2022-gemv2,lucy-etal-2024-aboutme}. 
This imbalance becomes even more striking at the cutting edge of AI, where generative models are rapidly expanding into multimodal territory~\citep{openai2024gpt4technicalreport, team2024gemini, TheC3, deitke2024molmopixmoopenweights, yue2025pangeafullyopenmultilingual, qwen2.5-VL}, seeking to represent a richer world made up of different modalities such as image, text, sound. 
In recent years, the community has made promising strides toward broader multilingual text evaluation~\citep{ahuja2023megamultilingualevaluationgenerative, singh-etal-2024-aya, singh2024global,aakanksha-etal-2024-multilingual,pozzobon2024manyexpandingscopetoxicity, romanou2024includeevaluatingmultilinguallanguage,singh2025globalmmluunderstandingaddressing,Adelani2024IrokoBenchAN}, and multimodal benchmarks are starting to take shape~\citep{pmlr-v162-bugliarello22a,fu2023mme,yue2023mmmu, yue2024mmmuprorobustmultidisciplinemultimodal,SEED-Bench10658180,xushao2025}.
Yet reliable evaluation at the intersection of multilingual and multimodal tasks remains rare. This gap is precisely what motivates our work.

One common but imperfect solution is translating English benchmarks into other languages. While convenient, this approach often falls short of capturing cultural context and nuance. Translated datasets can easily reinforce Western-centric knowledge and assumptions~\citep{van-miltenburg-etal-2017-cross,FRANK_ELLIOTT_SPECIA_2018,singh2025globalmmluunderstandingaddressing,longpre2025bridgingdataprovenancegap} limiting their ability to truly assess model performance across diverse settings. 
Moreover, automated data curation pipelines frequently amplify existing quality issues~\citep{luccioni-viviano-2021-whats, caswell-etal-2020-language, kreutzer-etal-2022-quality}, with translation artifacts such as \textit{translationese} muddying the waters even further~\citep{koppel-ordan-2011-translationese,zhang-toral-2019-effect,bizzoni-etal-2020-human,vanmassenhove-etal-2021-machine}.
While translated data has its place, especially for some particularly low-resource tasks~\citep{Zhou_2021_CVPR,thapliyal-etal-2022-crossmodal,qiu-etal-2022-multilingual,ramos-etal-2024-paella,geigle2025centuriodriversmultilingualability,dang-etal-2024-rlhf,ustun-etal-2024-aya,aakanksha-etal-2024-multilingual}, it is an imperfect substitute for genuinely diverse, in-language benchmarks.

In this work, we introduce the largest benchmark of real-world, in-language exam questions that blend image and text modalities. Our dataset pushes beyond simple captioning tasks, challenging models to reason about visual content in various topics, the way humans are evaluated in exams worldwide.
Through a large-scale open science effort across 18 languages, we construct~\benchmark (see Figure~\ref{fig:overview-figure}), featuring a diverse selection of knowledge domains across 14 subjects. 
With 55\% of the total 20,911 questions requiring image understanding for accurate resolution, our work aims to establish a comprehensive, and inclusive evaluation framework for multimodal language models. We evaluate a wide range of state-of-the-art models on~\benchmark, including Claude 3.5 Sonnet~\citep{TheC3}, GPT-4o~\citep{openai2024gpt4technicalreport}, and Gemini-V~\citep{team2024gemini}, as well as smaller open-weight VLMs, such as Aya-Vision model family~\citep{aya-vision}, Molmo~\citep{deitke2024molmopixmoopenweights}
Pangea~\citep{yue2025pangeafullyopenmultilingual}, and Qwen2.5-VL model family~\citep{qwen2.5-VL}.
Our key contributions and findings are highlighted here:

\begin{itemize}
    \item {\textbf{\textnormal{\benchmark} Benchmark}:} We present the largest multilingual multimodal exam set, covering high resource (e.g., English, Spanish) to underrepresented languages (e.g., Bengali, Telugu) across diverse subjects from sociology to STEM. Most languages (10/18) include 5+ topics, with the rest focusing on multi-subtopics like mathematics or engineering. Questions emphasize vision grounded reasoning through tasks like interpreting graphs, pictures, and region-specific diagrams, supported by fine-grained metadata for model diagnostics.
    \item \textbf{Modality-Specific Performance Disparities:} All models perform substantially better on text-only questions, revealing a clear disparity across modalities. The gap widens in larger modelsl; for instance, GPT-4o shows a 21.6\% difference between text-only and multimodal performance, while smaller models like Molmo exhibit a much narrower gap of 3.69\%. (Section~\ref{sec:overall_performance}). Furthermore, multimodal performance varies significantly by visual data type: models are more capable of answering questions about tables (76.5\%) and photographs (81.5\%) compared to diagrams (62.9\%).
    \item \textbf{Domain-Specific Performance Disparities:} We observe a significant performance gap between questions requiring knowledge of Humanities \& Social Sciences and those focused on STEM subjects (Section~\ref{sec:subject_results}). On average, models present accuracy of 83.7\% for humanities versus 59.2\% for STEM (based on the best scores across models). Models struggle more with STEM questions, suggesting that while they can often recognize visual content and retrieve related knowledge, they lack the reasoning capabilities needed to arrive at the correct answers in STEM domains.
    \item \textbf{Crosslingual Performance Disparities:} Model performance varies across languages, with noticeably better results in high-resource languages and weaker performance in mid- and low-resource ones (Section~\ref{sec:lang_results}). Crosslingual transfer appears to play a role, as models perform better on average in languages using Latin scripts compared to those with non-Latin scripts.
\end{itemize}

\section{The \textnormal{\benchmark} Benchmark}
The~\benchmark Benchmark is a global collection of multiple-choice questions sourced from real-world exams, with the goal of evaluating multimodal and multilingual understanding in VLMs. The collected exams are in a Multiple-choice question answering (MCQA) format which provides a structured framework for evaluation by prompting models with predefined answer choices~\citep{hendrycks2021measuring,lu2023mathvista,wang2024mathvision,yue2023mmmu,romero2024cvqa,romanou2024includeevaluatingmultilinguallanguage}, closely mimicking conventional human testing methodologies. %
Our work is built around three core design principles that guide the selection, curation, processing, and addition of exams: 
\begin{itemize}
    \item[\mmemoji] \textbf{Multimodality}: Images are central to \benchmark, as we aim to evaluate how VLMs integrate and reason about visual information to answer questions. We prioritize multimodal questions with diverse image types, complemented by a similar proportion of text-only questions for a complete assessment and comparison. 
    \item[\earthmoji\hspace{0.15em}] \textbf{Multilinguality}: The benchmark contains questions in 18 languages, with a focus on under-represented mid- and low-resource languages (e.g., Nepali, Lithuanian) alongside high-resource languages (e.g., English, Spanish) for a thorough evaluation across a broad range of languages. 
    \item[\diversitymoji] \textbf{Diversity}: Our goal is to collect exams covering as wide a range of topics as possible ranging from Mathematics and Sociology, to Medicine and Driving Licenses, ensuring comprehensive evaluation across various domains. 
    The final collection includes exams from 14 different domains, collected from 18  countries and with varying educational levels (from high school to professional exams), allowing detailed clustering and comprehensive evaluation.
\end{itemize}

\subsection{Global Collaboration}
Our work entailed an extensive, open science process to manually collect data by working directly with native speakers of different languages~\citep{elliott-etal-2016-multi30k,liu-etal-2021-visually,thapliyal-etal-2022-crossmodal,li-etal-2024-foodieqa, ustun-etal-2024-aya,singh-etal-2024-aya}. 
This is acutely needed in the field of machine learning, where recent studies have highlighted that dataset creators remain predominantly Western-centric~\citep{longpre2025bridgingdataprovenancegap}.  
The manual curation of datasets is a costly process that requires careful attention to detail in every language to ensure high-quality, contextually relevant content for evaluation. In this work, we engage in a large-scale open science collection process, which brings together contributors spanning 20 nations across four continents to ensure linguistic and cultural authenticity. For related participatory research see Section~\ref{sec:rel-work-particpatory}.

\subsection{Data Pipeline}
\textbf{Collection:} We collected \benchmark with guidelines detailing information about the type of exams and questions required, formatting, specifications, and quality control measures. The data was collected through a global call for contributions and distributed across global communities, with the majority of contributors being independent researchers in the Cohere for AI (C4AI)\footnote{\url{https://cohere.com/research}} open science community. 
This effort resulted in a collection of 20,911 questions from 18 different countries and 18 languages, all sourced in their original languages, avoiding translations to maintain linguistic authenticity. We prioritized original, domain-expert-written questions (e.g., from teachers), ensuring real-world relevance and quality. The exams were gathered from various repositories, including official government websites, question banks, and other publicly available repositories with educational materials. Throughout the process, contributors also annotated associated licenses with each dataset to allow for documentation of data provenance \citep{longpre2024}.

\begin{table*}[t!]
\centering
\resizebox{\textwidth}{!}{%
\begin{tabular}{llllllcc}
\toprule
\textbf{Language}   & \textbf{Code} & \textbf{Subjects} & \textbf{Total} & \textbf{Visual} & \textbf{Text} & \textbf{Resources}  & \textbf{Family} \\ 
\midrule
Portuguese & pt    & 11  & 2000 & 1000 & 1000 & High  & Italic \\
Serbian    & sr    & 1   & 2000 & 1000 & 1000 & High  & Balto-Slavic \\
Persian    & fa    & 5   & 2000 & 1000 & 1000 & High  & Iranian \\
Hindi      & hi    & 12  & 1886 & 1000 & 886  & High  & Indo-Aryan \\
Russian    & ru    & 1   & 1744 & 872  & 872  & High  & Balto-Slavic \\
English    & en    & 9   & 1628 & 814  & 814  & High  & Germanic \\
Spanish    & es    & 6   & 1482 & 741  & 741  & High  & Italic \\
Hungarian  & hu    & 1   & 1120 & 560  & 560  & High & Uralic \\
Dutch      & nl    & 10  & 1018 & 509  & 509  & High &  Germanic \\
French     & fr    & 1   & 762  & 381  & 381  & High &  Italic \\
German     & de    & 1   & 722  & 361  & 361  & High  & Germanic \\
Arabic     & ar    & 10  & 382  & 191  & 191  & High  & Semitic \\
Croatian   & hr    & 1   & 324  & 162  & 162  & High  & Balto-Slavic \\
Ukrainian  & uk    & 8   & 1237 & 1000 & 237  & Mid   & Balto-Slavic \\
Bengali    & bn    & 6   & 800  & 400  & 400  & Mid   & Indo-Aryan \\
Lithuanian & lt    & 6   & 680  & 340  & 340  & Mid   & Balto-Slavic \\
Telugu     & te    & 1   & 1000 & 1000 & 0    & Low   & South Dravidian \\
Nepali     & ne    & 1   & 126  & 126  & 0    & Low   & Indo-Aryan \\
\midrule
\textbf{Total} & (18) & 14 & 20,911 & 11,457 & 9,454 & -- & -- \\
\bottomrule
\end{tabular}%
}

\caption{\textbf{Statistics of the \textnormal{\benchmark} Dataset.}
Breakdown of subjects (Subjects), total questions (Total), multimodal questions (Visual), and text-only questions (Text) per language. Languages are covered by multiple sources with single-subject cases containing specialized subdomains. \mmemoji: Supports evaluation of both multimodal (image+text) and unimodal (text-only) capabilities.
\earthmoji\hspace{0.15em}: Languages are classified by resource level (high/mid/low) following~\citet{joshi-etal-2019-unsung, singh-etal-2024-aya}.
\diversitymoji: Enables granular analysis of model performance across modalities, languages, and subject domains.}
\label{table:dataset-statistics}
\end{table*}

\textbf{Processing:} The annotation process consists of two stages. In the first stage, we perform automated parsing and extraction. For directly parsable text, we use PDF or web parsers, while for non-parsable text, we employ OCR API's, such as Mathpix\footnote{\url{https://mathpix.com/convert}}, along with vision-language models such as GPT-4o. These tools allow us to extract both text and image elements from exam source formats, which are then converted into structured outputs in LaTeX, Markdown, and JSON formats, as required.
Since automated parsing can sometimes result in misaligned images and text, the second stage involves refining the extracted text. We apply heuristic rules, as well as high-performing LLMs, such as Claude 3.5 Sonnet and GPT-4o, to restructure the output, ensuring proper alignment of questions, text, and answer choices. This stage was followed by human verification, ensuring that images are correctly linked to the corresponding questions, and checking that extracted formulas match the expected equation format.

\textbf{Quality Assessment:}
\label{quality-assesment}
Maintaining reliable and high quality data is essential, especially given the large-scale international collaboration involved in this project. To ensure data integrity, we include manual validation in three stages of the collection and annotation pipeline. First, at the end of the collection stage, two independent annotators validate each exam to ensure conformity with the guidelines. Part of this verification includes a strict revision to confirm compliance with the distribution license requirements. Only exams approved by both independent annotators are included in the dataset. Next, following the annotation process, a validation script checks for JSON formatting errors, duplicate questions, and malformed strings that do not conform to identified entry specifications (see Appendix~\ref{dataset-fields}). Finally, at the last stage, two separate validators perform a final manual review of the collected files before merging them into \benchmark. 

Quality control extends to the evaluation phase as well, where we analyze the most prominent failure modes. During model inference, suspicious outputs, such as ambiguous answers, no response, or consistent failures across all models, are flagged for manual review. For example, if all models fail in a specific question, we investigate further and may discover issues like missing images or incorrect labels. If an issue is identified, the entire exam that contains the problematic question is reviewed for correction or removal. This process guarantees that any errors in the benchmark questions are identified and addressed, further enhancing the reliability of the dataset.

\subsection{Data Statistics}
The final \benchmark benchmark contains 20,911 questions across 18 languages belonging to 8 language families. A total of 11,459 questions require an image to be answered (55\%), while the remaining 9,452 (45\%) are text-only. The dataset covers 14 different subjects, grouped into 6 broad domains. Figure~\ref{fig:overview-figure} presents an overview of the dataset; detailed statistics can be found in Table~\ref{table:dataset-statistics}. 
The majority of questions in \benchmark are multimodal, with the exact proportion varying across languages, ranging from 50\% to 100\%, with some languages always requiring images for resolution.

Each exam question contains 17 fields, including source country, language, license, educational level, category, and multimodal information. These fields are detailed in Appendix~\ref{dataset-fields}. The questions are formatted in MCQA format with 4 options and a single correct answer. The subject is labeled in both English (\texttt{category\_en}) and the source language (\texttt{category\_source\_lang}). The educational level (e.g., high school, university entrance, professional licensing) is also included to ensure diverse representation. Multimodal questions additionally specify the type of image, such as graphs, tables, or diagrams. Additionally, each entry includes metadata such as source details, licensing status, and ISO 639-1 language codes.
For a fine-grained analysis, each question includes detailed metadata, with examples provided in Appendix~\ref{sec:visual-elements}. 
The metadata allows us to evaluate how visual and textual elements interact in multimodal reasoning tasks, making the benchmark valuable for evaluating models across diverse scenarios.

\benchmark covers a wide range of languages, including low- and mid-resource languages such as Nepali, Lithuanian, Bengali, Telugu, Persian, Ukrainian, Croatian, Serbian, and Hungarian, as well as high-resource languages such as English, Spanish, Portuguese, Russian, French, German, Arabic, Hindi, and Dutch. This selection allows us to evaluate how performance is affected by the amount of resources available for a given language. The dataset spans 8 different language families, providing a broad linguistic range. The number of questions per language varies significantly, from 126 for Nepali to 2000 for Portuguese, Serbian, and Persian. The linguistic diversity present in \benchmark enables a robust evaluation of models across both widely spoken and underrepresented languages, making the dataset suitable for comprehensive multilingual assessment.

\section{Experimental Setup}
\subsection{Models}
We benchmark both open-weights and closed multimodal vision-language models on~\benchmark, focusing on lighter open-weight models and larger closed models to assess performance across a wide range of model sizes. The open-weight models\footnote{All open-weight models are evaluated locally using 1$\times$NVIDIA Ampere A100 GPU with 64GB of memory for models up to 8B, and 4$\times$A100 for models on the range 32B--72B. To ensure a consistent evaluation environment, we set the temperature to 0.7, the maximum token generation to 1024, and the image size to 512×512 for all models.} include Aya-Vision-8B and 32B~\citep{aya-vision},
Molmo-7B-D~\citep{deitke2024molmopixmoopenweights}, Pangea-7B~\citep{yue2025pangeafullyopenmultilingual}, and all sizes of Qwen2.5-VL-Instruct~\citep{qwen2.5-VL} (3B, 7B, 32B, and 72B) to analyze the impact of model scale on~\benchmark. All models have image and multilingual support; Aya-Vision supports 23 languages, Qwen2.5-VL supports 29 languages, and Pangea was trained on a dataset spanning 39 different languages, making them strong candidates for multimodal and multilingual evaluation. For the closed models, we evaluate GPT-4o~\citep{openai2024gpt4technicalreport} (2024/08/06), Claude 3.5 Sonnet~\citep{TheC3} (2024/10/22), and Gemini 1.5 Pro~\citep{team2024gemini}.

\subsection{Evaluation Setup}
We designed two distinct evaluation setups to accommodate VLMs' varying reasoning and instruction-following capabilities.
For closed models, we designed zero-shot prompts using the Chain-of-Thought (CoT) method~\citep{wei2022chainofthought}. The prompt instructs the model to reason through the correct answer and to write the chosen option within specific \texttt{<ANSWER> </ANSWER>} tags. This approach is natural and aligns the real-world application of MCQs, encouraging the model to generate a step-by-step reasoning before selecting the final answer. We define a common template, ensuring equal evaluation conditions for all models (see Appendix~\ref{app:prompts}). The instructions were translated to all the evaluated languages, creating a fully in-language setup, following the methodology proposed by~\citet{romanou2024includeevaluatingmultilinguallanguage}. The selected choice is extracted using string matching of the tags.

For the smaller open-weight models, which typically have limited capacity for complex reasoning, CoT prompting proved less effective in our preliminary experiments (see Appendix~\ref{app:small-model-cot-results} for details). Therefore, we implemented a direct answer generation approach, instructing the models to produce a JSON output containing their choice within a predefined \texttt{`choice'} field. The instruction was always in English, independent of the question language (see Appendix~\ref{app:prompts}). This setup simplifies the task, reducing errors related to multi-step reasoning or formatting inconsistencies, and ensuring a straightforward answer extraction. Further discussion about model output error analysis can be found in~\cref{subsection:error-analysis}.

\subsection{Evaluation Metrics}
Given the multiple-choice nature of the task, we use accuracy as the primary evaluation metric. We report overall accuracy across all questions, as well as accuracy on the subset of questions where the model produces valid responses. A response is considered valid if the model successfully provides an answer in the expected format and selects a valid option (i.e., one of the letters {A, B, C, D}). Invalid responses typically result from missing the selected choice, selecting an invalid option, or refusal to answer. To quantify these cases, we report the \textit{Format Error Rate}, which measures the proportion of questions for which the model fails to generate a valid answer. For grouped results, we report the macro average across languages, i.e. all languages have equal weight when computing the score.

\section{Results}
\label{sec:overall_results}

\begin{table}[tb!]
\centering
\scriptsize %
\resizebox{\textwidth}{!}{
\begin{tabular}{@{}l*{9}{c}@{}}
\toprule
 & \multicolumn{3}{c}{\textbf{Overall}} & \multicolumn{3}{c}{\textbf{Multimodal}} & \multicolumn{3}{c}{\textbf{Text-only}} \\
\cmidrule(lr){2-4} \cmidrule(lr){5-7} \cmidrule(lr){8-10}
 & & \multicolumn{2}{c}{Valid Responses} & & \multicolumn{2}{c}{Valid Responses} & & \multicolumn{2}{c}{Valid Responses} \\ 
\cmidrule(lr){3-4} \cmidrule(lr){6-7} \cmidrule(lr){9-10} 
\textbf{Model} & Acc. & F.E. & Acc. & Acc. & F.E. & Acc. & Acc. & F.E. & Acc. \\
\midrule
Claude 3.5 Sonnet & \textbf{62.91} & 1.78 & \textbf{63.87} & \textbf{55.63} & 3.24 & \textbf{57.24} & \textbf{73.54} & 0.02 & \textbf{73.57} \\
Gemini 1.5 Pro & 62.10 & 1.62 & 62.95 & 55.01 & 1.46 & 55.71 & 72.35 & 1.81 & 73.45 \\
GPT-4o & 58.32 & 6.52 & 62.10 & 49.80 & 10.50 & 55.19 & 71.40 & 1.71 & 72.39 \\
\midrule
Qwen2.5-VL-72B & 52.94 & 0.02 & 53.00 & 48.40 & 0.03 & 48.41 & 60.00 & 0.02 & 60.01 \\
\hdashline
Aya-Vision-32B & 39.27 & 1.05 & 39.66 & 35.74 & 1.49 & 36.28 & 44.73 & 0.51 & 45.00 \\
Qwen2.5-VL-32B & 48.21 & 0.88 & 48.64 & 44.90 & 0.28 & 45.05 & 53.77 & 1.61 & 54.60 \\
\hdashline
Aya-Vision-8B & 35.09 & 0.07 & 35.11 & 32.35 & 0.05 & 32.36 & 39.27 & 0.10 & 39.30 \\
Molmo-7B-D & 32.87 & 0.04 & 32.88 & 31.43 & 0.06 & 31.44 & 35.12 & 0.01 & 35.13 \\
Pangea-7B & 31.31 & 7.42 & 34.02 & 27.15 & 13.52 & 31.02 & 37.84 & 0.03 & 37.86 \\
Qwen2.5-VL-7B & 39.56 & 0.08 & 39.60 & 36.85 & 0.04 & 36.88 & 43.91 & 0.11 & 43.96 \\
\hdashline
Qwen2.5-VL-3B & 35.56 & 0.19 & 35.63 & 33.67 & 0.32 & 33.79 & 38.51 & 0.03 & 38.53 \\
\bottomrule
\end{tabular}
}
\caption{
    \textbf{Performance Evaluation on \textnormal{\benchmark}}. 
    Results are reported as macro-averaged accuracy (\%) across all languages (equal weight per language). 
    \textbf{Acc.}: Accuracy over all samples; 
    \textbf{F.E.}: Format Error rate (invalid responses); 
    \textbf{Valid Acc.}: Accuracy excluding invalid responses. 
    Metrics are shown for the full dataset (\textbf{Overall}), multimodal inputs (\textbf{Multimodal}), and text-only inputs (\textbf{Text-only}).
    }
\label{tab:overall-performance}
\end{table}

\begin{figure}[ht!]
    \centering
    \captionsetup[subfigure]{justification=centering}
    \begin{subfigure}[b]{0.48\textwidth}
        \includegraphics[width=\textwidth]{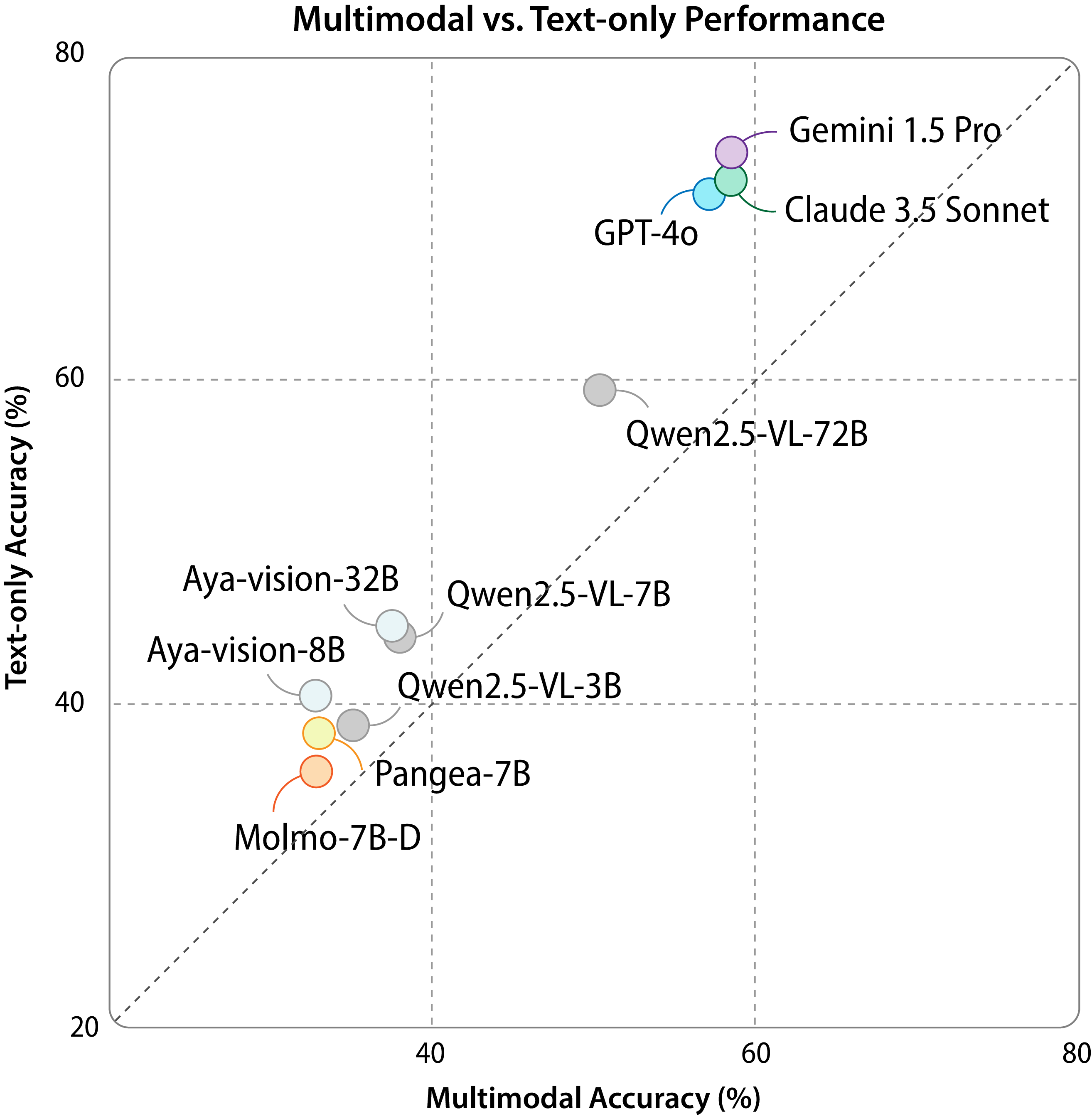} 
        \centering
        \caption{Accuracy as a function of modality.}
        \label{fig:multimodal-vs-text}
    \end{subfigure}
    \hfill 
    \begin{subfigure}[b]{0.48\textwidth}
        \includegraphics[width=\textwidth]{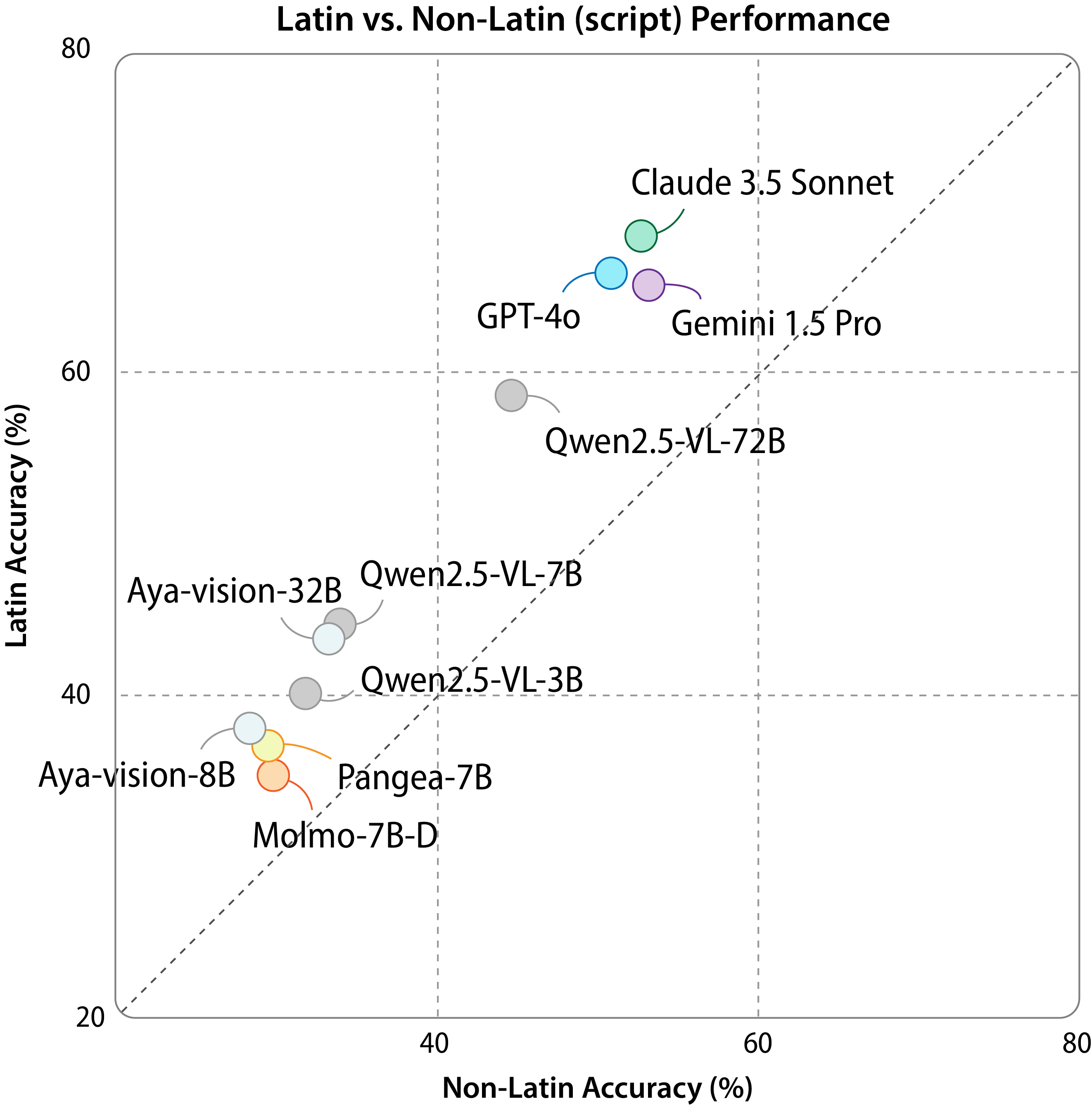} 
        \caption{Accuracy as a function of textual script.}
        \label{fig:latin-vs-nonlatin}
    \end{subfigure}
    \caption{\textbf{Model Performance Analysis on \textnormal{\benchmark}.}
(a) Accuracy (\%) of models on multimodal and text-only questions, highlighting low performance on multimodal samples.
(b) Accuracy (\%) by script type, revealing biases for latin scripts. Accuracy over valid responses is used to generate both figures. Identity line is added to show parity.
}
    \label{fig:dual-multimodal-script}
\end{figure}

\subsection{Overall Performance}
\label{sec:overall_performance}
We benchmark a wide variety of models on \benchmark and present the main results in Table~\ref{tab:overall-performance}.
Claude 3.5 Sonnet achieves the highest overall accuracy (62.91\%), followed closely by Gemini 1.5 Pro (62.10\%). GPT-4o performs notably worse (58.32\%), with a high format error rate (6.52\% overall, with at 10.50\% for the multimodal split). However, when considering only valid answers, GPT-4o’s performance improves significantly (+3.78 percentage points), closing the gap with other closed models and highlighting the impact of format errors (see Section~\ref{subsection:error-analysis}). 

Among open-weight models, Qwen2.5-VL-72B achieves the highest accuracy (52.94\%), which is expected given its larger number of parameters.  In the lightweight category ($\leq$8B parameters), Qwen2.5-VL-7B outperforms all others in both multimodal and text-only questions, with accuracy of 39.56\%. Open models generally maintain low format error rates, except for Pangea, which has the highest format error rate at 13\%. 

Table \ref{tab:overall-performance} also summarizes results for both the multimodal and text-only benchmark splits. 
Across all models, multimodal performance is lower than text-only, with a larger drop for closed models: GPT-4o drops 21.6 accuracy points overall (10.29 points for valid answers). Open-weight models show smaller gaps, with Molmo having the narrowest gap of only 3.69 accuracy points, though multimodal performance remains lower. Among lightweight models, Qwen2.5-VL-7B leads on both splits, with a relatively small gap, followed by Aya-Vision-8B on text-only samples and Qwen2.5-VL-3B in the multimodal split.
Closed models perform well on text-only questions, highlighting their strength in this modality but also the challenges of multimodal processing. In contrast, the smaller variation in performance across both splits by open-weight models suggests that they are less specialized, with lower overall performance but greater robustness in multimodal tasks. Lightweight models exhibit similar behavior across script types and modalities (Figure~\ref{fig:multimodal-vs-text}), with Molmo showing the most balanced performance between Multimodal vs. text-only samples and Latin vs. non-Latin scripts.

\subsection{Not All Image Types are Equal}

\begin{table}[ht!]
\centering
\resizebox{\textwidth}{!}{%
\begin{tabular}{@{}lcccccccc@{}}
\toprule
\textbf{Model} & 
\multicolumn{1}{c}{\textbf{Diagram}} & 
\multicolumn{1}{c}{\textbf{Figure}} & 
\multicolumn{1}{c}{\textbf{Graph}} & 
\multicolumn{1}{c}{\textbf{Map}} & 
\multicolumn{1}{c}{\textbf{Photo}} & 
\multicolumn{1}{c}{\textbf{Formula}} & 
\multicolumn{1}{c}{\textbf{Table}} & 
\multicolumn{1}{c}{\textbf{Text}} \\
& \footnotesize{(2,182)} & \footnotesize{(6,178)} & \footnotesize{(733)} & \footnotesize{(392)} & \footnotesize{(631)} & \footnotesize{(487)} & \footnotesize{(597)} & \footnotesize{(257)} \\
\midrule
Claude 3.5 Sonnet & \textbf{62.9} & 50.5 & \textbf{74.2} & \textbf{80.1} & 77.8 & 52.1 & 75.0 & 85.2\\
Gemini 1.5 Pro & 59.4 & \textbf{51.3} & 67.9 & 69.4 & 75.8 & \textbf{68.3} & 76.0 & 85.2 \\
GPT-4o & 59.6 & 48.2 & 68.4 & 78.8 & \textbf{81.5} & 64.4 & \textbf{76.5} & \textbf{86.2} \\
\midrule
Qwen2.5-VL-72B & 51.1 & 43.9 & 59.4 & 66.1 & 70.5 & 48.7 & 61.5 & 86.0 \\
\hdashline
Aya-Vision 32B & 38.6 & 33.4 & 42.0 & 50.0 & 60.2 & 32.4 & 33.1 & 68.8 \\
Qwen2.5-VL-32B & 46.7 & 41.0 & 53.1 & 58.2 & 65.0 & 47.3 & 58.0 & 82.5 \\
\hdashline
Aya-Vision 8B & 32.7 & 29.9 & 37.2 & 38.6 & 42.3 & 29.2 & 34.1 & 54.9 \\
Molmo-7B-D & 30.3 & 31.5 & 36.7 & 37.8 & 45.0 & 25.1 & 30.6 & 56.8 \\
Pangea-7B & 31.0 & 31.0 & 32.9 & 38.5 & 45.0 & 32.2 & 29.4 & 66.3 \\
Qwen2.5-VL-7B & 38.0 & 34.0 & 44.3 & 48.0 & 53.9 & 34.9 & 40.9 & 76.3 \\
\hdashline
Qwen2.5-VL-3B & 32.8 & 32.3 & 40.2 & 41.2 & 48.2 & 34.7 & 35.2 & 72.8 \\
\bottomrule
\end{tabular}%
}
\caption{\textbf{Model Performance Breakdown by Image Type in \textnormal{\benchmark}.} Accuracy (\%) over valid answers across image type. Bold values indicate top-performing model.
}
\label{tab:multimodal-imageType-acc}
\end{table}

\benchmark contains eight visual information types, with accuracy varying significantly by complexity (Table \ref{tab:multimodal-imageType-acc}). Simpler inputs like text-rich images (Qwen2.5-VL-7B: 76.3\%; GPT-4o: 86.2\%) and photos score higher than technical categories like Formulas and Diagrams (Qwen2.5-VL-7B: 38.0\%; GPT-4o: 62.9\%). Notably, Qwen2.5-VL-72B ranks second in text-rich images, surpassing both Gemini and Claude.
Larger models show specialized strengths: Gemini 1.5 Pro dominates Formulas and Figures, GPT-4o leads in text-rich images, and Claude 3.5 Sonnet achieves the highest scores in Diagrams, Graphs, and Maps. In contrast, Qwen2.5-VL-7B consistently outperforms all lightweight models across categories, demonstrating broader capability despite lower absolute scores. The results reveal a clear hierarchy: models handle simple visuals well but struggle with structured or symbolic data, a pattern consistent across architectures but more pronounced in smaller models. 

\subsection{Resource and Script Sensitivity in Models}\label{sec:lang_results}

\begin{figure}
    \centering
    \includegraphics[width=1\linewidth]{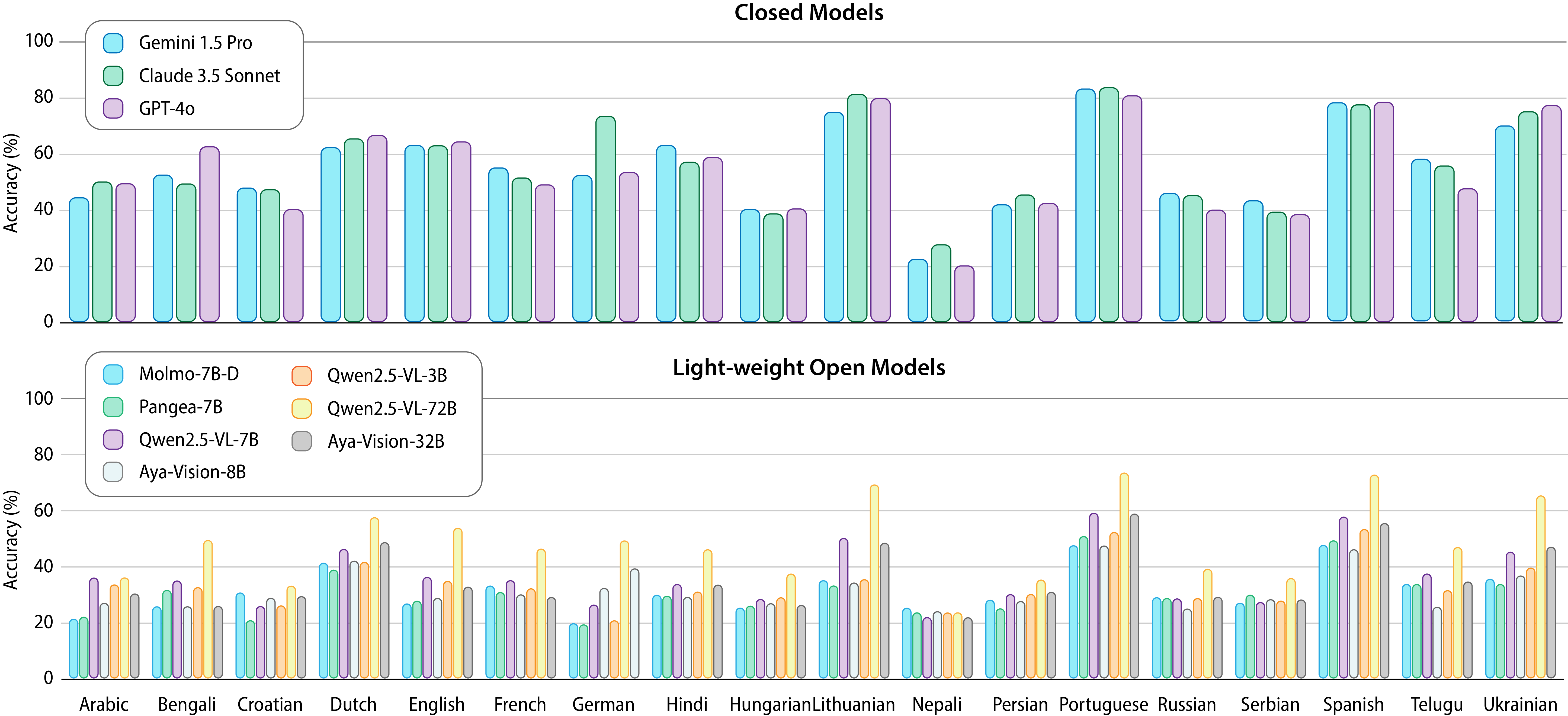}
    \caption{\textbf{Multimodal Accuracy by Language in \textnormal{\benchmark}.} Reports performance (accuracy \%) for closed models and open-weight models on multimodal questions.}
    \label{fig:multimodal-lang-acc}
\end{figure}

Performance in \benchmark varies widely across all 18 languages  (see Figure~\ref{fig:multimodal-lang-acc}). Models generally perform well in high-resource languages (e.g., English, Spanish, German) but struggle with lower- and mid-resource ones, such as Nepali and Telugu. This can be attributed to the limited training data for these languages, complex scripts, and the exclusive use of multimodal samples for these languages (see Table~\ref{table:dataset-statistics}), which are inherently more challenging. 
Lithuanian, despite being mid-resource language, stands out as the highest-performing language, with Claude 3.5 Sonnet leading in overall accuracy. 
This might be due the fact that all Lithuanian questions belong to \texttt{College Graduation Exams}, and have a major subject composition of Social Sciences and Humanities in opposition to STEM subjects, which may align well with the models’ capabilities. 
Closed models show similar performance within each language, except for German, where Claude excels. In contrast, Qwen2.5-VL-7B consistently leads all lightweight models for almost every language, and the heavier Qwen2.5-VL-72B shows the benefits of model scale.

The results show that all models are biased towards Latin script languages. As shown in Figure~\ref{fig:latin-vs-nonlatin}, all models are above the parity line, exhibiting consistent higher performance for Latin scripts compared to non-Latin scripts. Full results can be found in Table~\ref{tab:complete-lang-acc}.

\subsection{STEM Questions Expose Model Deficiencies}\label{sec:subject_results}

\begin{table}[htb!]
\centering
\resizebox{0.8\textwidth}{!}{%
\begin{tabular}{@{}lccccccccccc@{}}
\toprule
                & \multicolumn{3}{c}{Closed Weights}            & \multicolumn{7}{c}{Open Weights} \\ \cmidrule(lr){2-4} \cmidrule(lr){5-11}
 &
  \multicolumn{1}{c}{\rotatebox{0}{\makecell{
  \rotatebox{90}{Gemini}}}} &
  \multicolumn{1}{l}{\rotatebox{0}{\makecell{\rotatebox{90}{{Claude}}}}} &
  \multicolumn{1}{c}{\rotatebox{0}{\makecell{\rotatebox{90}{{GPT-4o}}}}} &
  \multicolumn{1}{c}{\rotatebox{0}{\makecell{\rotatebox{90}{{Qwen2.5-3B}}}}} &
  \multicolumn{1}{c}{\rotatebox{0}{\makecell{\rotatebox{90}{{Molmo-7B}}}}} &
  \multicolumn{1}{c}{\rotatebox{0}{\makecell{\rotatebox{90}{{Pangea-7B}}}}} &
  \multicolumn{1}{c}{\rotatebox{0}{\makecell{\rotatebox{90}{{Qwen2.5-7B}}}}} &
  \multicolumn{1}{c}{\rotatebox{0}{\makecell{\rotatebox{90}{{Aya V-8B}}}}} &
  \multicolumn{1}{c}{\rotatebox{0}{\makecell{\rotatebox{90}{{Aya V-32B}}}}} &
  \multicolumn{1}{c}{\rotatebox{0}{\makecell{\rotatebox{90}{{Qwen2.5-72B}}}}} \\ \midrule
\multicolumn{11}{l}{\hspace{-1ex}\textit{Humanities \& Social Sciences}}\\\midrule
Economics       & 64.1          & 63.8          & \textbf{66.7} & 37.7 & 27.5          & 33.9 & 42.7 & 30.9 & 29.8 & 58.8 \\
Geography       & 72.8          & \textbf{81.5} & 80.4          & 40.7 & 37.6          & 36.7 & 51.0 & 39.5 & 50.5 & 70.4 \\
History         & 78.7          & 83.7          & \textbf{86.4} & 48.9 & 42.1          & 42.4 & 52.9 & 45.6 & 61.4 & 77.1 \\
Language        & 83.5          & 85.5          & \textbf{85.8} & 72.2 & 60.1          & 66.0 & 75.7 & 56.6 & 71.2 & 85.1 \\
Social Sciences & 85.7          & 82.9          & \textbf{88.1} & 52.9 & 52.2          & 53.8 & 68.6 & 58.0 & 64.3 & 80.0 \\
Sociology       & 92.3          & \textbf{93.4} & 93.2          & 64.1 & 61.0          & 57.3 & 73.1 & 57.7 & 70.5 & 87.2 \\
\midrule
\multicolumn{11}{l}{\hspace{-1ex}\textit{STEM}}\\\midrule
Biology         & 60.3          & 62.9          & \textbf{63.9} & 37.6 & 35.3          & 33.4 & 42.6 & 35.4 & 40.7 & 53.8 \\
Chemistry       & \textbf{60.4} & 59.7          & 52.9          & 33.2 & 33.5          & 34.1 & 38.5 & 28.0 & 34.8 & 50.0 \\
Engineering     & 57.3          & \textbf{64.4} & 56.3          & 28.9 & 24.4          & 24.2 & 32.4 & 30.3 & 34.8 & 48.4 \\
Mathematics     & \textbf{48.6} & 44.4          & 44.0          & 30.4 & 28.8          & 29.0 & 30.1 & 28.6 & 29.6 & 40.3 \\
Physics         & 57.8          & \textbf{58.7} & 54.7          & 33.7 & 26.7          & 28.9 & 34.7 & 27.1 & 33.0 & 42.3 \\
\midrule
\multicolumn{11}{l}{\hspace{-1ex}\textit{Reasoning, Health Science, and Practical Skills}}\\\midrule
Reasoning       & 52.0          & \textbf{53.3} & 51.0          & 27.4 & 27.5          & 26.6 & 29.5 & 25.1 & 27.6 & 42.3 \\
Medicine        & 70.2          & 73.8          & \textbf{75.6} & 36.7 & 40.4          & 38.4 & 45.8 & 35.4 & 52.3 & 63.3 \\
Driving License & 64.4          & 64.2          & \textbf{73.1} & 39.0 & 44.9 & 39.4 & 44.9 & 41.6 & 47.1 & 54.5 \\
\bottomrule
\end{tabular}
}
\caption{\textbf{Subject-wise Performance on \textnormal{\benchmark}'s Multimodal Questions.} Valid accuracy (\%) across examination subjects for multimodal samples, with bold highlighting top-performing models.}
\label{tab:multimodal-cat-acc}
\end{table}

\benchmark consists of exams covering 14 subjects and domains, with Table~\ref{tab:multimodal-cat-acc} summarizing model performance on the multimodal split across subjects. We observe that all models perform significantly better on Humanities \& Social Science questions compared to other domains. The closed models achieve high accuracy in areas like Sociology (Claude: 93.4\%, GPT-4o: 93.2\%), Social Sciences (GPT-4o: 88.1\%, Gemini: 85.7\%), and Language (GPT-4o: 85.8\%, Claude: 85.5\%). In contrast, performance in STEM subjects, including Mathematics, Physics, and Engineering, is notably lower, with most models scoring below 50\%. This suggests that while they are generally capable of recognizing visual content and retrieving surface-level knowledge, they fall short when it comes to performing the multi-step reasoning and problem-solving required in STEM subjects. Answering these questions often demands not just factual recall but also the ability to interpret complex diagrams, apply mathematical concepts, and reason through scientific principles \---\ capabilities that current models have yet to fully master. This highlights a key gap in their ability to bridge perception and reasoning, particularly in tasks that require deeper analytical thinking.

\section{Analysis}
\label{subsection:error-analysis}

\subsection{How Sensitive Are VLMs to Missing or Incorrect Images?}

\begin{table}[hbt]
\small
\centering
\begin{tabular}{lccc}
\toprule
& & \multicolumn{2}{c}{\textbf{Valid Responses}} \\
\cmidrule{3-4}
\textbf{Setup}               & \textbf{Accuracy} & \textbf{Format Error} & \textbf{Accuracy} \\ \midrule
Standard Multimodal & 36.85                    &0.04                                   & 36.88                                        \\
Random Image        & 32.56                     & 3.12                                   & 33.53                                   \\ 
No Image               & 33.44                     & 0.03                                   & 33.45                                       \\ \bottomrule
\end{tabular}
\caption{\textbf{Image Relevance Analysis for Qwen2.5-VL-7B on \textnormal{\benchmark}.} Model performance across the standard multimodal, \textit{Random Image}, and \textit{No-Image} setups to assess the impact of visual information on question-answering accuracy.}
\label{tab:qwen-noimage-results}
\end{table}

To evaluate the dependency of multimodal questions on images, and the impact of incorrect image associations, we conducted an experiment using the multimodal split of \benchmark. Following~\citet{elliott-2018-adversarial,thomason-etal-2019-shifting}, we created two modified versions of the dataset: (1) a \textit{`No Image’} split, where all images were removed, and (2) a \textit{`Random Image’} split, where images were randomly reassigned to questions. The aim of this experiment is to assess how much the models rely on the visual information. We evaluate the performance of Qwen2.5-VL-7B on these modified splits, and the results are shown in Table \ref{tab:qwen-noimage-results}. 

We observe that the model performs above the random baseline (25\%) across all three splits, indicating some ability to reason from text alone. However, there is a drop in performance ($-$3.41\% in Total Accuracy) when questions are presented without images, suggesting that the model does rely on visual information for accurate answers. The performance drop is similar for both modifications; however, we observe a significantly larger format error when the model is tested with irrelevant images. In several of these cases, the model actually acknowledges that the image does not correspond to the question. In contrast, in experiments with no images, the format error rate is almost zero, indicating that the model attempts to answer even when visual inputs are missing.\footnote{We observed that Qwen2.5-VL-7B tends to hallucinate when no image is present. In a simple experiment using the prompt \texttt{``Describe the following image''}, the model correctly describes the input image when provided. However, when no image is passed, the model hallucinates and generates a random description.
}

\subsection{Scaling Model Size Improves Performance}

\begin{figure}
    \centering
    \includegraphics[width=0.8\linewidth]{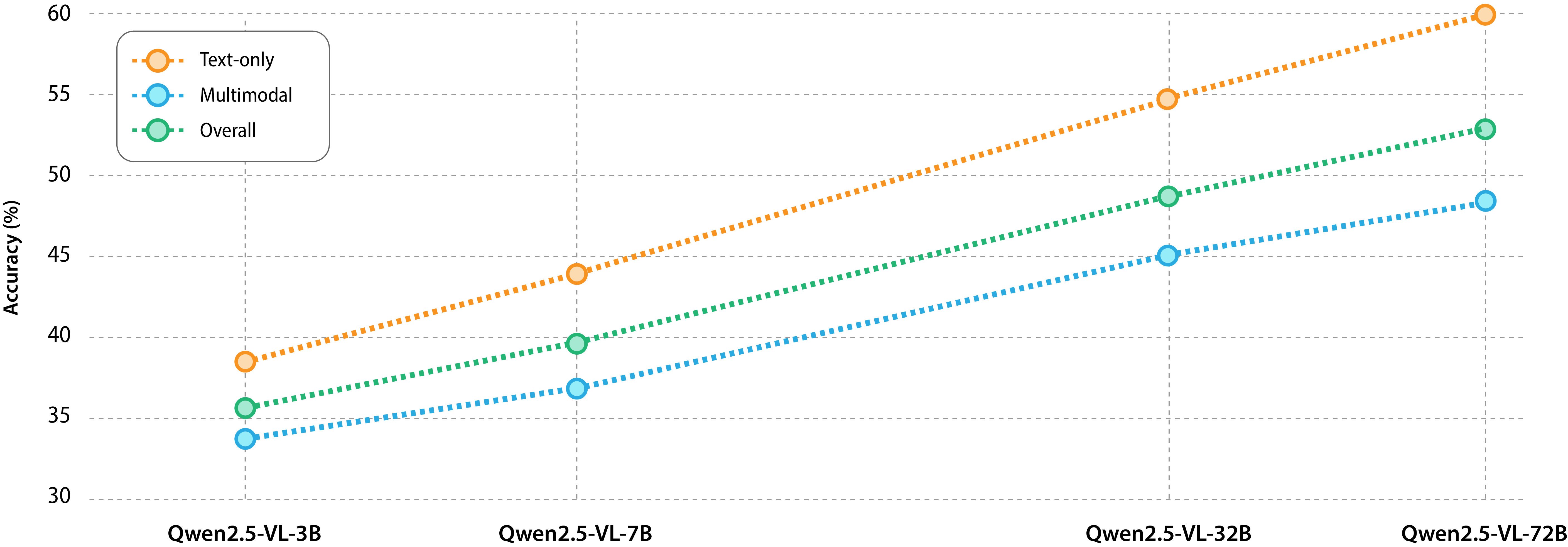}
    \caption{\textbf{Model Size Analysis for Qwen2.5-VL Models.} Performance improvement across three model sizes (3B, 7B, 32B, and 72B parameters) on \benchmark's multimodal tasks, demonstrating consistent gains from increased model capacity. Note that x-axis is shown in log-scale.
    }
    \label{fig:scaling}
\end{figure}

To analyze the impact of model size on \benchmark performance, we evaluated all four variants of Qwen2.5-VL. We selected this model family for its well-distributed size range, as well as being the best performing model in the open weight model category. We follow the same experimental setup for all model versions. 

Figure~\ref{fig:scaling} shows the performance of Qwen2.5-VL variants on \benchmark. Model size is shown in the x-axis (log-scale), while the y-axis displays accuracy for multimodal and text-only splits, and overall score. We observe a linear relationship between the logarithm of the model size and accuracy, with larger models showing significant gains. The largest open model model evaluated, Qwen2.5-VL-72B, still underperforms the closed models, however, these results highlight the effectiveness of scaling for open models, with clear and predictable improvements at each size tier.

\subsection{To What Extent Do Textual Augmentations Boost VLM Capabilities?}
\label{subsubsection:captioning-results}

The significant performance gap between text-only and multimodal responses raises critical questions about the strengths and weaknesses of the visual processing in the tested models. In this analysis, we investigate to what extent do visual processing constraints limit multimodal capabilities, and conversely, can automatically generated textual augmentation improve model performance? 

To explore this direction, we generate synthetic captions (using Gemini 1.5 Pro) and Optical Character Recognition (OCR) text (Tesseract~\citep{tesseract}) for all images in \benchmark, aligning with the methodology of ~\citep{das-2024-examsv}. Unlike prior work that completely replaces images with text, we evaluate whether a VLM \emph{augmented} with these textual inputs can boost performance. 

\begin{table}[ht]
\small
\centering
\begin{tabular}{lccccc}
\toprule
        & & \multicolumn{2}{c}{Qwen2.5-VL-7B}          & \multicolumn{2}{c}{Gemini 1.5 Pro}              \\
        \cmidrule(lr){3-4} \cmidrule(lr){5-6}
        & Samples & Image & +Caption & Image & +Caption \\
       \midrule
Diagram & 2,182 & \textbf{38.0}             & 37.9           & 59.4       & \textbf{59.6}                      \\
Figure & 6,178 & 34.0                      & \textbf{34.8}           & \textbf{51.3}       & 50.0                               \\
Graph & 733  & 44.3                      & \textbf{45.2}           & 67.9       & \textbf{68.2}                      \\
Map & 392    & \textbf{48.0}             & 46.7           & 69.4       & \textbf{70.9}                      \\
Photo & 631  & 53.9                      & \textbf{54.1}           & \textbf{75.8}       & 74.3                               \\
Formula & 487 & 34.9                      & \textbf{37.3}           & 68.3       & \textbf{68.7}                      \\
Table & 597 & \textbf{40.9}             & 34.6           & 76.0       & \textbf{76.1}                      \\
Text & 257 & 76.3                      & \textbf{79.8}           & \textbf{85.2}       & 83.7                               \\ \toprule
Macro Avg. & 11,457 & \textbf{36.88} &  36.83 & \textbf{55.71} & 54.81 \\
\bottomrule 
\end{tabular}

\caption{\textbf{Accuracy on augmented multimodal inputs with image captions.} Results are grouped by image type. We report \textbf{Valid Accuracy (\%)}; the highest scores are highlighted in bold for each model. Macro averaged accuracy is reported over language for both methods.
}
\label{tab:caption-imtype}
\end{table}

Table~\ref{tab:caption-imtype} shows the results of augmenting visual inputs with synthetic captions and OCR text across diverse image types in \benchmark, measured by valid accuracy (\%). Overall, the addition of a caption and OCR text improves the performance of the selected models in 5 out of 8 image types. Both models experienced a performance boost coordinately for \textit{Graph} and \textit{Formula}. The experiment reveals that the utility of textual augmentation depends critically on image content type. While Gemini 1.5 Pro dominates overall performance, Qwen2.5-VL-7B demonstrates selective gains when provided with captions and OCR: improvements in \textit{Graph} (+0.9\%), \textit{Photo} (+0.2\%), \textit{Formula} (+2.4\%), and \textit{Text} (+3.5\%) suggest that textual augmentation aids interpretation of content where visual elements are tightly coupled with symbolic or linguistic features (e.g., labeled axes, embedded text, or mathematical notation). Conversely, performance declines for \textit{Diagram} ($-0.1$\%), \textit{Map} ($-1.3$\%), and \textit{Table} ($-6.3$\%) with augmentation, implying that synthetic captions may introduce noise or fail to capture structural relationships critical to these categories. Gemini’s robustness across modalities ($\leq2\%$ variation in most categories) suggests its stronger native visual understanding reduces reliance on supplementary text. The results underscore that captioning effectiveness is context-dependent: text augmentation benefits models most when (1) visual content inherently contains extractable text (e.g., \textit{Photo} with signs, \textit{Text} regions) or (2) symbolic patterns (e.g., formulas, graphs) require disambiguation. However, for structurally complex or text-sparse images (e.g., \textit{Map}, \textit{Diagram}), captioning may not compensate for deficiencies in spatial or relational reasoning. Full results, including total accuracy and format error, can be found in Table~\ref{tab:complete-captioning}.

\subsection{Format Errors}
\label{subsection:format-error-analysis}

\begin{figure}
    \centering
    \includegraphics[width=1\linewidth]{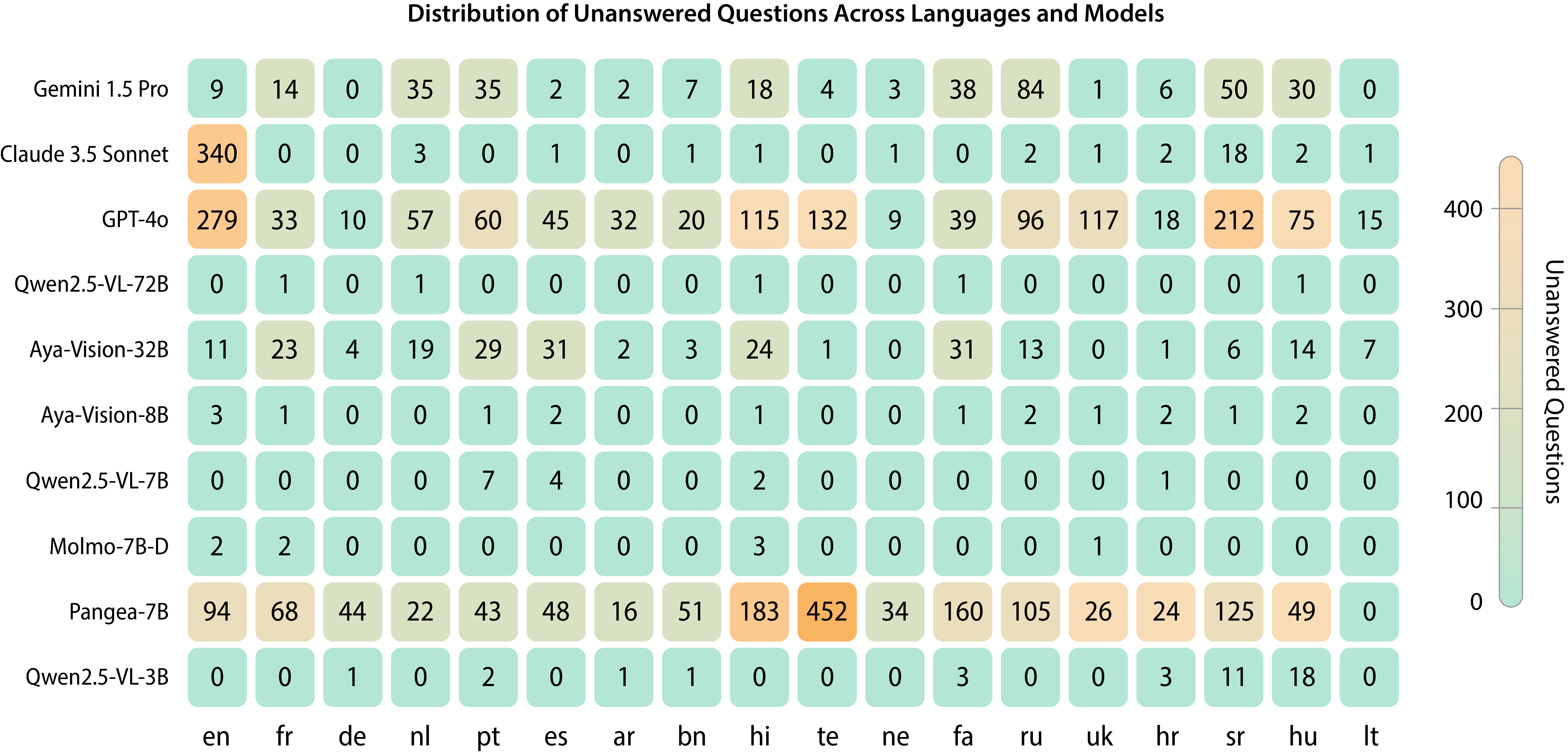}
    \caption{\textbf{Distribution of the number of format errors for each model/language combination}. The languages are represented in their ISO 639 (set 1) code.}
    \label{fig:heatmap-unanswered-by-lang}
\end{figure}

While our experimental setup ensures a majority of answers were extracted from model outputs, we observe occasional failures: models struggle to follow instructions, the outputs contain formatting errors, or models refuse to answer (particularly for health-related or ethical questions).
Figure~\ref{fig:heatmap-unanswered-by-lang} shows that unanswered questions concentrate in mid- to low-resource languages, and the distribution accumulates over non-latin scripts, likely due to tokenization challenges, insufficient language-specific training data, or  visual-textual alignment difficulties. Pangea-7B shows the highest refusal rates, especially for Telugu (452), Hindi (130), Persian (160), and Serbian (125). While other open models show minimal unanswered counts, indicating better format adherence.  Closed models (Claude 3.5 Sonnet, GPT-4o) display distinct behavior: their refusals concentrate on non-Latin, low-resource languages, but they also show high error rates for English questions, primarily health/medical queries due to policy constraints. This underscores the trade-off between content moderation and benchmark performance.

\section{Related Work}
\subsection{General Challenges in Multilingual Evaluation of Vision-Language Models}

VLMs have demonstrated impressive performance in processing and generating text, interpreting images, and reasoning across multiple modalities. These advances have been driven by various multimodal benchmarks~\citep{li2024survey, Vayani2024AllLM, nayak-etal-2024-benchmarking, schneider2025gimmickgloballyinclusive} that assess capabilities such as image captioning, object attribute recognition, and spatial relationship understanding. However, most existing benchmarks prioritize high-resource languages (e.g., English~\citep{zang2024contextual, schneider2025gimmickgloballyinclusive} or Chinese~\citep{fu2023mme, he2024cmmu}), resulting in significant gaps in multilingual evaluation. This focus creates disparities, particularly in evaluating performance on low-resource non-Latin languages~\citep{hengle2024multilingualneedlehaystackinvestigating}. A common strategy for lower-resource languages has been to translate existing English benchmarks using tools such as ChatGPT~\citep{lai-etal-2023-okapi}, GPT-4~\citep{yue2025pangeafullyopenmultilingual}, or Google Translate~\citep{li2023bactrian}. Such approaches, however, often fail to capture the linguistic and cultural diversity necessary for global applications~\citep{singh2024global, huang2025surveylargelanguagemodels}, may introduce errors or employ uncommon terms, thereby affecting the reliability of assessments, and exacerbate the problems with poverty-conscious language technology~\citep{bird-2022-local}. Furthermore, the lack of comprehensive evaluation suites for non-English languages has substantially hindered multilingual generative advancements, limiting LLMs' ability to perform equitably across diverse linguistic landscapes, a challenge particularly critical for evaluating toxicity and biases in multilingual settings~\citep{ustun-etal-2024-aya}. Addressing these limitations is essential for ensuring that VLMs perform robustly across a diverse range of real-world scenarios.

Recent efforts have attempted to address these shortcomings by incorporating culturally and linguistically diverse data. Only recently have we seen some multilingual benchmarks. For example, in the reasoning space, MMLU-ProX~\citep{mmlu-prox} has created a comprehensive reasoning benchmark in 13 languages. Culturally-diverse Multilingual Visual Question Answering Benchmark (CVQA)~\citep{romero2024cvqa} creates culturally relevant multiple-choice questions about images from 30 countries in 31 languages, using local languages which are then translated into English. In contrast, \benchmark builds on this work by nearly doubling the number of questions while focusing on 18 languages, thereby allowing for a more in-depth evaluation of each language. Moreover, \benchmark focuses on multiple-choice exams covering a wide range of topics beyond culturally relevant MCQA, including also STEM exams scenarios. Additional work in evaluating VLMs includes PangeaBench, a holistic evaluation suite encompassing 14 datasets (13 pre-existing) split between multimodal and text-only tasks, covering 47 languages~\citep{yue2025pangeafullyopenmultilingual}. Similarly,~\citet{Vayani2024AllLM} introduce a multimodal benchmark that includes culturally diverse images paired with text across 100 languages. Notably, this benchmark incorporates non-MCQA formats (e.g., True/False and free-form answers), which is a key distinction from the MCQA format adopted in \benchmark.

Other benchmarks further illustrate the diversity of evaluation approaches. For instance, MaRVL~\citep{liu-etal-2021-visually} assesses images in binary framework, which limits possible nuances in cultural evaluations. CULTURALVQA~\citep{nayak-etal-2024-benchmarking} emphasizes cultural knowledge with approximately 44.1\% of its data focusing on rituals and traditions; however, it relies on open-ended questions in English, which contrasts with the MCQA and multilingual approach of \benchmark. Moreover, the MaXM benchmark~\citep{MaXM} addresses bias and provides multilingual, multimodal assessment across 7 languages but does not focus on cultural aspects, an area where \benchmark offers added value.

\subsection{Exam-Style Benchmarks for Vision-Language Models}

Exam-style benchmarks have also advanced the evaluation of VLMs (see Table~\ref{app:compare-benchmarks}).~\citet{zhang2023m3exam} present M3Exam, a novel benchmark sourced from real human exam questions that tests models in a multilingual, multimodal, and multilevel context using an MCQA framework. M3Exam includes 12,317 questions in 9 languages, requiring both multilingual proficiency and cultural knowledge; however, only about 23\% of its questions necessitate image processing. This benchmark highlighted challenges faced by state-of-the-art models, such as GPT-4, particularly in handling low-resource and non-Latin script languages alongside complex multimodal queries. In contrast, \benchmark differentiates itself by covering a larger number of languages.~\citet{das-2024-examsv} present EXAMS-V, a multi-discipline, multimodal, multilingual exam benchmark comprising 20,932 multiple-choice questions across 20 school disciplines (spanning natural sciences, social sciences, religion, fine arts, and business). EXAMS-V includes questions in 11 languages from 7 language families and incorporates four categories of multimodal features (scientific symbols, figures, graphs, and tabular data). Despite this, only 5,086 of its questions are multimodal. Additionally, the M5 benchmark~\citep{schneider-sitaram-2024-m5} evaluates VLMs on diverse vision-language tasks in a multilingual and multicultural context by covering 41 languages across eight datasets and five tasks, though it does not utilize an MCQA format.
In contrast, \benchmark stands out from these existing exam-based benchmarks by featuring a broader diversity of languages and the largest proportion of multimodal questions in an MCQA format, with 55\% of its questions requiring image understanding. This makes \benchmark, a comprehensive and challenging multimodal and multilingual testing ground for evaluating vision-language models in multilingual real-scenarios.

\subsection{Participatory Open Science Projects}\label{sec:rel-work-particpatory}
Participatory research empowers diverse communities to actively contribute to research processes, capturing linguistic subtleties and cultural nuances directly from native speakers. Prior participatory NLP research has primarily targeted region-specific tasks such as translation, character recognition, and audio transcription. We highlight notable initiatives here which served as our motivation and backbone framework for building \benchmark.

In Africa, the \textbf{Masakhane}\footnote{\url{https://www.masakhane.io/}} community exemplifies impactful participatory NLP by focusing on grassroots-led data collection, annotation, and model creation for African languages.~\citet{nekoto-etal-2020-participatory} demonstrated that communities in low-resource environments significantly contribute to NLP, even without formal training. Subsequent efforts by~\citet{adelani-etal-2023-masakhanews} have further advanced dataset curation and model development for underrepresented African languages using similar participatory frameworks. Similarly, the \textbf{MaRVL} dataset (Multicultural Reasoning over Vision and Language;~\citealp{liu-etal-2021-visually}) employed native speakers from diverse linguistic backgrounds (\textit{Indonesian, Swahili, Tamil, Turkish}, and \textit{Mandarin Chinese}) to contribute culturally representative images, subsequently annotated by professional linguists. Despite its cultural richness, MaRVL's modest scale (under 8,000 data points) limits broader applicability beyond evaluation.

In Latin America, participatory research has also emerged and is continuously growing through the help of communities. Recent works include ~\citet{hernandez-mena-meza-ruiz-2022-creating}, which developed eight open-access linguistic resources via structured social service programs, engaging student volunteers in transcription and segmentation tasks. Concurrently,~\citet{CaneteCFP2020} and \citet{guevara-rukoz-etal-2020-crowdsourcing} spearheaded crowd-sourced corpora addressing dialectal diversity and resource scarcity specific to Latin American Spanish. 

In Southeast Asia, \textbf{Project SEALD}\footnote{\textbf{S}outh\textbf{e}ast \textbf{A}sian \textbf{L}anguages in One Network \textbf{D}ata; \url{https://aisingapore.org/aiproducts/southeast-asian-languages-in-one-network-data-seald/}}, a collaboration between AI Singapore and Google Research, facilitated multilingual dataset collection to support regional Large Language Models (LLMs). Outputs from SEALD underpin open-source multilingual models such as \textit{SEA-LION}\footnote{\url{https://sea-lion.ai}}, \textit{WangchanLion}~\citep{phatthiyaphaibun2024wangchanlionwangchanxmrceval}, and \textit{Sahabat-AI}\footnote{\url{https://sahabat-ai.com}}. Related initiatives include \textbf{NusaCrowd} for aggregating and standardizing Indonesian NLP datasets~\citep{cahyawijaya-etal-2023-nusacrowd} and the \textbf{SEACrowd} and \textbf{SEA-VL} projects aimed at comprehensive evaluation and benchmarking of LLMs across Southeast Asian languages~\citep{cahyawijaya2025crowdsource,lovenia-etal-2024-seacrowd}.

On a global scale, the \textbf{CVQA dataset}~\citep{romero2024cvqa} was created using a participatory approach, involving native speakers and cultural experts from over 30 countries. Annotators were selected for their fluency in local languages and cultural familiarity. Many contributors were also recognized as co-authors based on their level of involvement, reinforcing a collaborative, community-driven effort. The \textbf{Aya Initiative} employed participatory methods, engaging over 3,000 contributors to curate instruction datasets across 114 languages, resulting in one of the largest multilingual datasets for language model training~\citep{singh-etal-2024-aya, ustun-etal-2024-aya}. Similarly, the INCLUDE benchmark~\citep{romanou2024includeevaluatingmultilinguallanguage} leveraged participatory approaches closely aligned with our methodology. The \textbf{BigScience ROOTS corpus}, developed collaboratively for the BLOOM model, exemplifies large-scale participatory data collection. Approximately 62\% of ROOTS data was crowd-sourced via global hackathons and open submissions, involving over 1,000 researchers from 60 countries and more than 250 institutions, resulting in 1.6 terabytes of multilingual data~\citep{NEURIPS2022_roots}.
Additionally,~\citet{Uzuner_10.1136/jamia.2010.004200} underscored the viability of community-driven annotation for complex, domain-specific NLP tasks like clinical text annotation, highlighting broader applicability of participatory frameworks beyond general NLP domains.

Participatory methods have also successfully extended into reinforcement learning from human feedback (RLHF). For instance, the \textbf{OpenAssistant} project, led by LAION, utilized global crowdsourcing to construct a multilingual corpus comprising over 161,000 messages annotated by 13,500 volunteers. This dataset facilitated robust training of dialogue-aligned language models through extensive human feedback annotations~\citep{Kopf_NEURIPS2023_949f0f8f}.

\section{Conclusion}
\label{sec:conclusion}

As generative models become increasingly multimodal and multilingual, the need for robust and culturally grounded evaluation benchmarks has never been more urgent. In this work, we take a step toward closing this gap by introducing the largest benchmark of real-world, in-language multimodal exam questions. By grounding evaluation in authentic exam settings from around the world, our benchmark challenges models to reason about images in ways that mirror human assessment, capturing both linguistic and cultural complexity.

Our findings highlight the limitations of current models in handling this intersection of skills: multilingual understanding, visual reasoning, and culturally aware problem-solving. We hope this benchmark serves not only as a valuable tool for measuring progress but also as a call to action for developing models that are truly capable of operating across languages, cultures, and modalities.
Continued investment in representative, high-quality evaluation datasets will be essential to ensure that future AI systems are equitable and globally relevant.

\section*{Limitations}

While our benchmark represents an important step toward more representative multilingual multimodal evaluations, several limitations still remain.
First, the dataset is inherently imbalanced across languages. Coverage varies depending on the availability and accessibility of exam sources, with some languages significantly underrepresented.
Second, difficulty levels are not uniformly controlled. Since questions are drawn directly from real-world exams across diverse educational systems, variations in exam design, curricular focus, and intended grade levels introduce potential inconsistency in task complexity across languages and modalities. Further the chosen MCQA question format, inherent to many exams, has issues, see Appendix~\ref{app:mcqaframework}. For instance: \textbf{Exploitation of biases:} Models may guess correct answers by exploiting statistical patterns or poorly designed distractors, inflating performance metrics without demonstrating genuine understanding. \textbf{Limited real-world applicability:} Unlike open-ended queries typical in real-world applications, MCQA provides predefined options, which may not reflect natural user interactions. \textbf{Choice-order sensitivity:} Performance can vary based on the order of answer choices, introducing inconsistencies unrelated to model capability.
Finally, while the dataset expands coverage beyond English, the overall language diversity remains limited. Many languages, especially those spoken in low-resource regions, are still missing due to the scarcity of suitable exam material and annotators. 

\section*{Acknowledgments}

We acknowledge the EuroHPC Joint Undertaking for awarding this project access to the EuroHPC supercomputer LEONARDO, hosted by CINECA (Italy) and the LEONARDO consortium through an EuroHPC Development Access call (ID:EUHPC\_D12\_071). This work was supported by research grant (VIL53122) from Villum Fonden, and by the European Union’s Horizon 2020 research and innovation program under grant agreement No.
101135671 (TrustLLM). EF gratefully acknowledges the support of the Googler Initiated Grants and the Google Award for Inclusion Research programs. MZ is supported by the research grant (VIL57392) from Villum Fonden. JMI is supported by the National
University Philippines and the UKRI Centre for Doctoral Training in Accountable, Responsible, and Transparent AI [EP/S023437/1] of the University of Bath.

\bibliography{custom, anthology}
\clearpage
\appendix

\section{Data Collection Details}
\label{sec:appendix}

\subsection{Dataset Fields}
\label{dataset-fields}
\begin{table*}[ht]
\centering
\begin{tabular}{p{0.27\linewidth}  p{0.7\linewidth}}
\toprule
\textbf{Field} & \textbf{Description} \\
\midrule
\texttt{language} & The language in which the question is written (e.g., \texttt{"en"} for English). \\
\midrule
\texttt{country} & The country where the exam originated (e.g., \texttt{"United States"}). \\
\midrule
\texttt{contributor\_country} & The contributor's country of residence (e.g., \texttt{"Spain"}). \\
\midrule
\texttt{file\_name} & The internal database filename for the original exam document. \\
\midrule
\texttt{source} & The URL or reference to the original exam document. \\
\midrule
\texttt{license} & Licensing information of the exam (e.g., \texttt{"Unknown"} if not stated). \\
\midrule
\texttt{level} & The educational level of the exam (e.g., \texttt{"University Entrance"}). \\
\midrule
\texttt{category\_en} & The exam subject category in English (e.g., \texttt{"Chemistry"}). \\
\midrule
\texttt{category\_source\_lang} & The subject category as written in the original language (\texttt{language}). \\
\midrule
\texttt{original\_question\_num} & The original question number in the source document. \\
\midrule
\texttt{question} & The text of the question. \\
\midrule
\texttt{options} & A list of possible answer choices. \newline
    For example, $[$\texttt{"Option A"}, \texttt{"Option B"}, \texttt{"Option C"}, \texttt{"Option D"}$]$. \\
\midrule
\texttt{answer} & The index of the correct answer  (e.g., \texttt{3} for the fourth option). \\
\midrule
\texttt{question\_image} & The extracted diagram, graph, or table associated with the \texttt{question}. \\
\midrule
\texttt{image\_information} & A label indicating the importance of the \texttt{question\_image} for answering the question. Possible values include: \\
                        & \tabitem \texttt{"useful"} - The image provides additional clarification. \\
                        & \tabitem \texttt{"essential"} - The image is necessary to answer the question. \\
\midrule
\texttt{image\_type} & The category of \texttt{question\_image} (e.g., \texttt{"figure"}, \texttt{"graph"}, \texttt{"table"}) as described in Appendix~\ref{sec:visual-elements}.\\
\bottomrule
\end{tabular}
\caption{Structured dataset fields with descriptions used in data collection protocol.}
\label{tab:ds_fields}
\end{table*}

\subsection{Categories of Visual Elements}
\label{sec:visual-elements}

We group the visual elements into eight primary categories in \benchmark.  If an image falls into multiple categories, we assign the most representative based on the image's content.

\includepdf[pages=-]{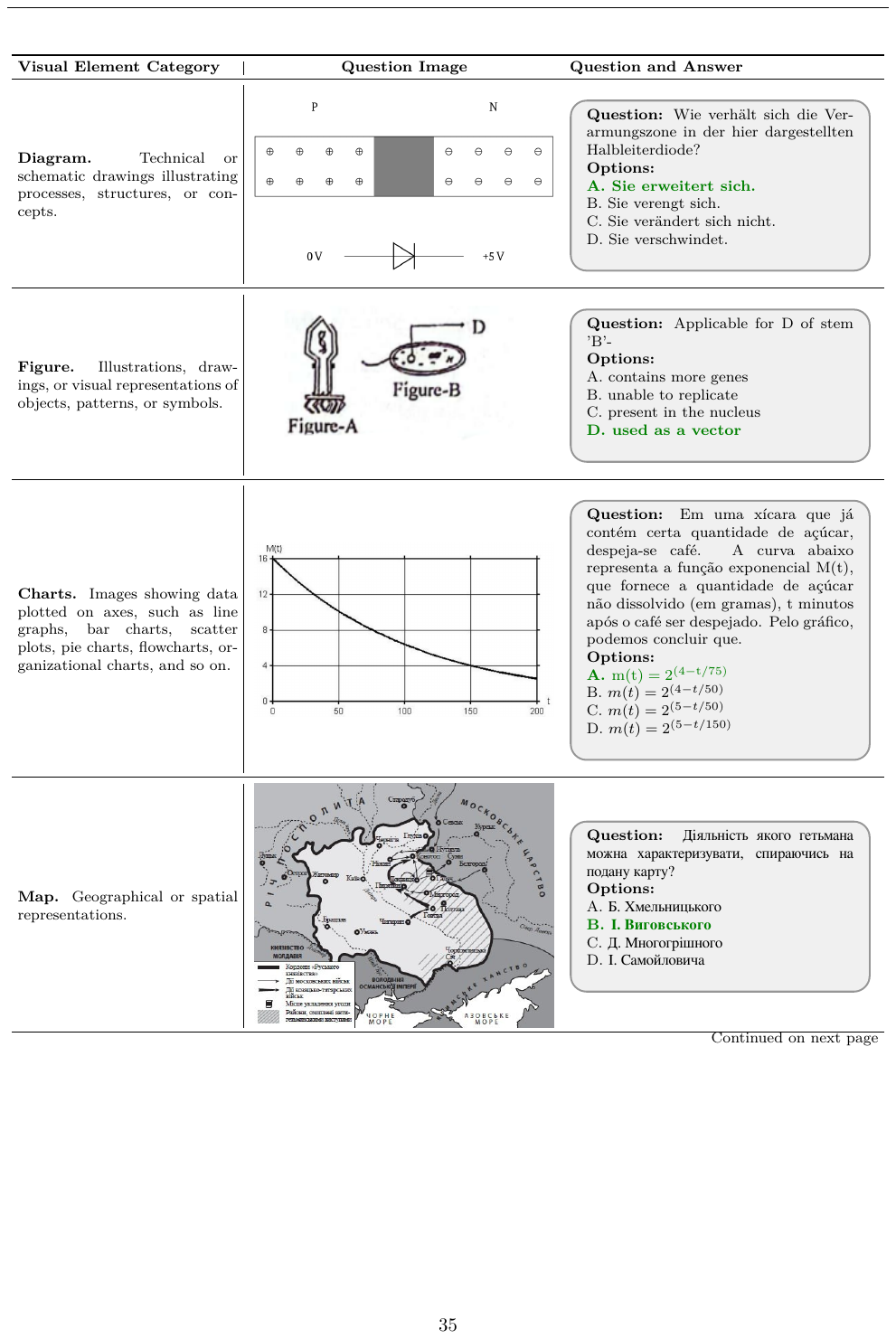}

\subsection{Complete Results}
We report full multimodal performances in table~\ref{tab:complete-lang-acc} for each language and in table~\ref{tab:complete-cat-acc} for each subject. Each table reports, for each model and category; \textbf{Total Accuracy \%}: the accuracy over all samples, \textbf{Valid Accuracy \%}: the accuracy over successfully extracted answers and \textbf{Format Error \% (FE)}: the proportion of unextracted answers. 
\label{sec:complete-results}

\setcounter{table}{8}
\input{tables/complete_lang_acc}

\input{tables/complete_cat_acc}

We report full multimodal performances in table~\ref{tab:complete-lang-acc} for each language and in table~\ref{tab:complete-cat-acc} for each subject. Each table reports, for each model and category; \textbf{Total Accuracy \%}: the accuracy over all samples, \textbf{Valid Accuracy \%}: the accuracy over successfully extracted answers and \textbf{Format Error \% (FE)}: the proportion of unextracted answers. 

\input{tables/complete_captioning}

\subsection{License}
To ensure ethical data usage, we prioritize sources that permit redistribution and academic use. During data collection, we filter out content from sources with restrictive licensing policies. Additionally, our dataset does not include personally identifiable information, and all collected exams are either publicly available or obtained under appropriate agreements.

\subsection{Prompts}
\label{app:prompts}
The prompts that we used to perform all experiments were designed to ensure consistency across languages. Examples are shown both in English and Spanish as an overview. Below is a summary of the key components.

\subsubsection{System Message}
\label{subsubsec:system-messages}
A system message sets the context for the model, instructing it to act as an expert in solving multiple-choice questions. For zero-shot CoT prompting, the message is provided in all the evaluation languages to support language-specific evaluation.
\begin{itemize}
    \item \textbf{Zero-shot CoT}:
    
        \begin{itemize}
            \item English: \texttt{You are an expert at solving multiple-choice questions. Carefully analyze the question, think step by step, and provide your FINAL answer between the tags <ANSWER> X </ANSWER>, where X is ONLY the correct choice. Do not write any additional text between the tags.}
            \item Spanish: \texttt{Eres un experto en resolver preguntas de opción múltiple. Analiza cuidadosamente la pregunta, piensa paso a paso y proporciona tu respuesta FINAL entre las etiquetas <ANSWER> X </ANSWER>, donde X es ÚNICAMENTE la opción correcta. No escribas ningún texto adicional entre las etiquetas.}
        \end{itemize}

    \item \textbf{Direct answer}: 

    \texttt{You are a helpful assistant who answers multiple-choice questions. For each question, output your final answer in JSON format with the following structure: \{"choice": "The correct option (e.g., A, B, C, or D)"\}. ONLY output this format exactly. Do not include any additional text or explanations outside the JSON structure.} 

\end{itemize}

\subsubsection{Keywords}
\label{subsubsec:keywords}
Language-specific keywords are used to structure the prompts consistently across languages. These include terms for "Question," "Options," and "Answer" to be included when generating the prompt. For example:
\begin{itemize}
    \item English: \texttt{\{"question": "Question", "options": "Options", "answer": "Answer"\}}
    \item Spanish: \texttt{\{"question": "Pregunta", "options": "Opciones", "answer": "Respuesta"\}}
\end{itemize}

\subsubsection{Prompt Examples}
System messages~\ref{subsubsec:system-messages} and Keywords~\ref{subsubsec:keywords} are used to systematically craft the prompt for a model in a specific language. We show examples of both a closed and an open model in Table~\ref{tab:prompt_examples}.

\definecolor{modelA}{RGB}{64,81,181}
\definecolor{modelB}{RGB}{171,71,188}

\begin{table}[H]
  \centering
  \begin{tabular}{>{\raggedright\arraybackslash}m{0.45\linewidth} >{\raggedright\arraybackslash}m{0.45\linewidth}}
    \toprule
    \textbf{\textbf{Open Model}} & \textbf{\textbf{Closed Model}} \\
    \midrule
    \begin{minipage}[t]{\linewidth}
      \footnotesize  
      \textcolor{modelA}{\texttt{SYSTEM:}}\\
      \textcolor{modelA}{\texttt{You are a helpful assistant who answers multiple-choice questions. For each question, output your final answer in JSON format with the following structure: \detokenize{"choice":"The correct option (e.g., A, B, C, or D)"}. ONLY output this format exactly. Do not include any additional text or explanations outside the JSON structure. Output your choice in the specified JSON format.}}\\

      \textcolor{modelA}{\textbf{\texttt{USER:}}}\\[0.5ex]
      \includegraphics[width=0.8\linewidth]{}\\[1ex]
      \textcolor{modelA}{\texttt{Question: Make CORRECT match between Group-I and Group-II, in relation to}}\\
      \textcolor{modelA}{\texttt{interaction between two species.}}\\
      \textcolor{modelA}{\texttt{Options:}}\\
      \textcolor{modelA}{\texttt{A.) P-I, Q-II, R-III, S-IV}}\\
      \textcolor{modelA}{\texttt{B.) P-III, Q-II, R-IV, S-I}}\\
      \textcolor{modelA}{\texttt{C.) P-IV, Q-III, R-II, S-I}}\\
      \textcolor{modelA}{\texttt{D.) P-III, Q-IV, R-II, S-I}}\\
      \textcolor{modelA}{\texttt{Answer:}}
    \end{minipage}
    &
    \begin{minipage}[t]{\linewidth}
      \footnotesize  
      \textcolor{modelB}{\textbf{\texttt{SYSTEM:}}}\\
      \textcolor{modelB}{\texttt{Eres un experto en resolver preguntas de opción múltiple. Analiza cuidadosamente la pregunta, piensa paso a paso y proporciona tu respuesta FINAL entre las etiquetas <ANSWER> X </ANSWER>, donde X es ÚNICAMENTE la opción correcta. No escribas ningún texto adicional entre las etiquetas.}}\\
      \textcolor{modelB}{\textbf{\texttt{USER:}}}\\[0.5ex]
      \includegraphics[width=0.8\linewidth]{}\\[1ex]
      \textcolor{modelB}{\texttt{Pregunta: Ante esta imagen en un paciente con un trastorno motor en miembros inferiores, señale la respuesta INCORRECTA:}}\\
      \textcolor{modelB}{\texttt{Opciones:}}\\
      \textcolor{modelB}{\texttt{A.) Debemos buscar una malformación de Chiari}}\\
      \textcolor{modelB}{\texttt{B.) En algunos casos se asocia a hidrocefalia}}\\
      \textcolor{modelB}{\texttt{C.) Se caracteriza por una pérdida de la sensibilidad táctil y vibratoria con preservación de la sensación térmica y dolorosa}}\\
      \textcolor{modelB}{\texttt{D.) Puede producirse tras traumatismos o infecciones}}\\[1ex]
      \textcolor{modelB}{\texttt{Respuesta:}}
    \end{minipage}\\[1ex]
    \bottomrule
  \end{tabular}
  \centering
  \caption{\textbf{Prompt examples in \textnormal{\benchmark}.} Multimodal prompt samples with interleaved image are shown for an open model and a closed model.}
  \label{tab:prompt_examples}
\end{table}

\subsection{Captioning \& OCR}
We instantiated Gemini 1.5 Pro with the following instruction to generate synthetic captions from the images in \benchmark. Prompts with image augmentations are shown in Table~\ref{tab:caption_prompt_examples}.
\begin{table}[H]
  \centering
  \begin{tabular}{>{\raggedright\arraybackslash}m{0.45\linewidth} >{\raggedright\arraybackslash}m{0.45\linewidth}}
    \toprule
    \textbf{\textbf{Open Model}} & \textbf{\textbf{Closed Model}} \\
    \midrule
    \begin{minipage}[t]{\linewidth}
      \footnotesize  
      \textcolor{modelA}{\texttt{SYSTEM:}}\\
      \textcolor{modelA}{\texttt{You are a helpful assistant who answers multiple-choice questions. For each question, output your final answer in JSON format with the following structure: \detokenize{"choice":"The correct option (e.g., A, B, C, or D)"}. ONLY output this format exactly. Do not include any additional text or explanations outside the JSON structure.}}\\
    
      \textcolor{modelA}{\textbf{\texttt{USER:}}}\\[0.5ex]
      \includegraphics[width=0.8\linewidth]{}\\[1ex]

      \textcolor{green}{\texttt{Caption: }}\textcolor{modelA}{\texttt{\detokenize{The code initializes character variables 'a' to 'P' and 'b' to 'x'. It then calculates 'c', 'd', and 'e' using bitwise operations (&, |, ^) and character addition with '*', '-', and '+', respectively.  The `printf` function outputs the characters c, d, and e. Below the code, three tables display ASCII values:  one for uppercase letters 'A' to 'Z' (65 to 90), another for lowercase letters 'a' to 'z' (97 to 122), and a third for symbols '*', '+', and '-' (42, 43, and 45, respectively). Ellipses (...) within the tables indicate omitted values between the shown characters.}}}
      \textcolor{green}{\texttt{OCR: }}\textcolor{modelA}{\texttt{\detokenize{#include<stdio.h>\}\n\nint main(int argc,\n\nchar a = 'P';\nchar b = 'x';\nchar c = (a &\nchar d = (a |\nchar e = (a \u201c\n\nprintf (\"sc \%\nreturn 0;\n\n\}\n\nchar *argv[]) \{\n\nby + te;\nb) - '-\%3\nb) + \"Hy\n se\\n\", c, d, e);\n\nASCII encoding for relevant characters is given below\n\n42| 43) 45\n\n}}}\\
      \textcolor{modelA}{\texttt{Question: What is printed by the following ANSI C program?}}
      \textcolor{modelA}{\texttt{Options:}}
      \textcolor{modelA}{\texttt{A.) z K s}}
      \textcolor{modelA}{\texttt{B.) 122 75 83}}
      \textcolor{modelA}{\texttt{C.) * - +}}
      \textcolor{modelA}{\texttt{D.) P x +}}\\
      \textcolor{modelA}{\texttt{Answer:}}
    \end{minipage}
    &
    \begin{minipage}[t]{\linewidth}
      \footnotesize  
      \textcolor{modelB}{\textbf{\texttt{SYSTEM:}}}\\
      \textcolor{modelB}{\texttt{Eres un experto en resolver preguntas de opción múltiple. Analiza cuidadosamente la pregunta, piensa paso a paso y proporciona tu respuesta FINAL entre las etiquetas <ANSWER> X </ANSWER>, donde X es ÚNICAMENTE la opción correcta. No escribas ningún texto adicional entre las etiquetas.}}\\ \\
      \textcolor{modelB}{\textbf{\texttt{USER:}}}\\[0.5ex]
      \includegraphics[width=0.5\linewidth]{}\\[1ex]
      \textcolor{green}{\texttt{Caption: }}\textcolor{modelB}{\texttt{This table presents data on total weight, measured in grams, across four months.  The table consists of two columns: \"Mes\" (Month) and \"Peso total\" (Total Weight). Month 1 shows a weight of 1,500 grams, Month 2 shows 2,600 grams, Month 3 shows 3,700 grams, and Month 4 shows 4,800 grams. The table is a simple grid format with plain black text on a white background.}}\\
      \textcolor{green}{\texttt{OCR: }}\textcolor{modelB}{\texttt{\detokenize{Wes | Pesototal\n1 [1500 gramos\n2 |_2600 grmos\n3 | 3700 gemoe\na\n\n\u201cZOO grams\n\n}}}\\
      \textcolor{modelB}{\texttt{Pregunta: Un perro cachorro tenía un peso de 1.500 gramos al mes de nacido. En la tabla se muestra el peso del cachorro en los primeros cuatro meses. De acuerdo con la tabla, ¿Cuál es el cambio del peso del cachorro entre un mes y el mes siguiente?}}\\
      \textcolor{modelB}{\texttt{Opciones:}}\\
      \textcolor{modelB}{\texttt{A.) Disminuyó 1.500 gramos}}\\
      \textcolor{modelB}{\texttt{B.) Disminuyó 2.600 gramos}}\\
      \textcolor{modelB}{\texttt{C.) Aumentó 1.100 gramos}}\\
      \textcolor{modelB}{\texttt{D.) Aumentó 3.300 gramos}}\\
      \textcolor{modelB}{\texttt{Respuesta:}}
    \end{minipage}\\[1ex]
    \bottomrule
  \end{tabular}
  \centering
  \caption{\textbf{Caption+OCR prompt examples in \textnormal{\benchmark}.} Prompts are shown for open and closed models in English and Spanish. Caption and OCR additions are highlighted in green.}
  \label{tab:caption_prompt_examples}
\end{table}

\textbf{Gemini 1.5 Pro's prompt for captioning:}
\begin{verbatim}
**Instruction:**
You are an expert image captioner. Generate highly detailed, precise, 
and academically relevant textual descriptions of images sourced from exam questions, 
ensuring all critical visual elements are captured for accurate problem-solving.

**Guidelines:**

Exam-Specific Analysis:

- Primary Elements: Identify and describe key components (e.g., diagrams, charts, graphs,
labels, symbols, annotations) and their exact attributes (e.g., numerical values, units,
directional arrows, text annotations).

- Secondary Details: Note stylistic features (e.g., "black-and-white schematic," 
"color-coded bars in a graph"), spatial relationships (e.g., "force vectors 
pointing northwest"), and contextual clues (e.g., axes labels, legends, scales).

- Textual Elements: Explicitly transcribe all visible text (e.g., labels like 
"Mitochondria," numbers like "5V," titles like "Figure 2: Velocity vs. Time").

Academic Precision:

- Technical Focus: Prioritize details critical to exam questions (e.g., "a right 
triangle with hypotenuse labeled c = 10 cm," "a bar graph comparing GDP of 5 countries,
with Japan’s bar shaded blue at 4.3 trillion").

- Diagrams/Charts: Specify type (e.g., "pie chart," "circuit diagram") and components 
(e.g., "resistor symbol connected to a battery").

- Scientific Relevance: Highlight measurements, units, symbols (e.g., "T = 25°C," 
"a pulley system with frictionless ropes").

Structure & Clarity:

- Begin with the image’s purpose (e.g., "A biology diagram of a plant cell") 
followed by a systematic breakdown (left-to-right, top-to-bottom, or by functional layers).

- Use neutral, objective language. Avoid assumptions unless implied by context (e.g.,
"a downward arrow labeled 9.8 m/s² likely representing gravitational acceleration").

**Output Format:**

- Single paragraph (4–6 sentences).

- Example:
"A physics diagram depicts two blocks on a frictionless inclined plane: Block A (5 kg)
is connected via a rope to Block B (3 kg) over a pulley. Angle theta = 30°, with 
vectors labeled F_normal and F_gravity. A scale beside the plane shows time t = 0s
to t = 5s. Text at the bottom reads: ‘Calculate tension in the rope.’ The image is 
monochrome, with dashed lines indicating motion direction."

Constraints:

- Avoid Omissions: Ensure no labels, numbers, or symbols are overlooked, even if 
small or peripheral.

- Neutral Tone: Exclude subjective interpretations (e.g., "messy handwriting" 
or "complex diagram") unless style is exam-relevant (e.g., "a hand-drawn sketch with 
annotations").
\end{verbatim}

\subsection{Open-Weight Models CoT Results}
\label{app:small-model-cot-results}
To benchmark the models, we initially designed a CoT prompt that instructed the models to think step-by-step and then provide the correct answer, marking the choice with the tags \texttt{<ANSWER>} \texttt{</ANSWER>}. However, in preliminary experiments, we found this instruction too complex for mid- to small-sized models (32B–3B), which struggled to follow it consistently.

In Table~\ref{tab:results-cot-vs-direct}, we compare results using CoT versus the direct English-language prompt adopted in our final evaluation. The error rate was considerably higher for most models, even after cleaning and extracting answers with regex matching their typical output formats. Two exceptions were Pangea and Molmo, which showed lower error rates with the CoT prompt. However, this was because both ignored the reasoning instruction and simply output the selected option, making extraction easier. Prompt choice significantly impacted performance: the direct English prompt improved results across all models except Pangea, which remained unchanged.

\begin{table}[H]
\centering
\scriptsize %
\resizebox{0.9\textwidth}{!}{
\begin{tabular}{@{}l*{6}{c}@{}}
\toprule
 & \multicolumn{3}{c}{\textbf{Overall CoT}} & \multicolumn{3}{c}{\textbf{Overall In-English}} \\
\cmidrule(lr){2-4} \cmidrule(lr){5-7} 
 & & \multicolumn{2}{c}{Valid Responses} & & \multicolumn{2}{c}{Valid Responses} \\ 
\cmidrule(lr){3-4} \cmidrule(lr){6-7} 
\textbf{Model} & Acc. & F.E. & Valid Acc. & Acc. & F.E. & Valid Acc. \\
\midrule
Aya-Vision-32B & 38.94 & 8.04 & 42.06 & 39.27 & 1.05 & 39.66 \\
\hdashline
Aya-Vision-8B & 33.08 & 6.22 & 35.15 & 35.09 & 0.07 & 35.11 \\
Molmo-7B-D & 32.86 & 0.01 & 32.87 & 32.87 & 0.04 & 32.88  \\
Pangea-7B & 31.24 & 5.61 & 33.45 & 31.31 & 7.42 & 34.02 \\
Qwen2.5-VL-7B & 35.18 & 6.34 & 37.64 & 39.56 & 0.08 & 39.60 \\
\hdashline
Qwen2.5-VL-3B & 32.90 & 1.40 & 33.33 & 35.56 & 0.19 & 35.63 \\
\bottomrule
\end{tabular}
}
\caption{
    \textbf{Comparison of CoT and direct English prompting on \textnormal{\benchmark} for small models.}. 
    Reported values are macro-averaged accuracy (\%) across all languages.
    }
\label{tab:results-cot-vs-direct}
\end{table}

\subsection{Comparison with Other Benchmarks} 
\label{app:compare-benchmarks}
Table~\ref{app:compare-benchmarks} offers a concise comparison of key multimodal benchmarks. MMMU~\citep{yue2023mmmu}, SEED-Bench~\citep{SEED-Bench10658180}, and MME~\citep{fu2023mme} are single-language datasets focused mainly on image-text pairs, with SEED-Bench also incorporating video-text. MME is notably smaller and only partially human-annotated, using mostly true/false formats. In contrast, M3Exam \citep{zhang2023m3exam}, EXAMS-V~\citep{das-2024-examsv}, and M5~\citep{schneider-sitaram-2024-m5} introduce multilingualism—M5 being the most extensive with 41 languages—though much of its content is not multiple-choice and lacks verified annotations.

\begin{table*}[h]
\centering
\scriptsize
\renewcommand{\arraystretch}{1.2}
\begin{tabularx}{\textwidth}{p{2.5cm} X X X X X X}
\toprule
\textbf{Benchmark} & \textbf{Languages} & \textbf{Samples}& \textbf{Multimodal} & \textbf{Modalities} & \textbf{Human Annotation} & \textbf{Answer type}\\ \midrule
\textbf{MMMU}~\citep{yue2023mmmu} & 1 & 11,550 & 11,264 & Image-Text & Yes  & MCQA    \\ 
\textbf{SEED-Bench}~\citep{SEED-Bench10658180} & 1 & 19,242 & 19,242 & Image-Text, Video-Text & Partial  & MCQA \\  %
\textbf{MME}~\citep{fu2023mme} & 1$^\dag$ & 2,194 & 0  & Image-Text &  Partial & Y/N  \\ 
\textbf{M3Exam}~\citep{zhang2023m3exam} & 9 & 12,317 & 2,816  & Image-Text  & Yes & MCQA \\ 
\textbf{EXAMS-V}~\citep{das-2024-examsv}& 11 & 20,932 & 5,086  & Image-Text &  Yes  & MCQA   \\ 
\textbf{M5}~\citep{schneider-sitaram-2024-m5} & 41 & 237,094 & 1,422 & Image-Text & Yes  &  Mix  \\
\hdashline
\textbf{\benchmark} & \textbf{18} & \textbf{20,911} & \textbf{11,459} & Image-Text & Yes & MCQA \\ 
\bottomrule
\end{tabularx}

\caption{\textbf{Comparison of Multimodal Benchmarks.} $^\dag$All but 40 questions are in English that measure machine translation capability from Chinese to English.
}
\label{tab:multimodal_benchmarks}
\end{table*}

\benchmark stands out by offering a balanced composition of 20,911 samples across 18 languages, with a strong focus on multimodal reasoning (11,459 Image-Text samples), comprehensive human annotation, and a consistent multiple-choice setup. Compared to existing benchmarks, \benchmark is more linguistically diverse than M3Exam and EXAMS-V, includes more multimodal samples than M5, and ensures higher quality through expert-verified annotations, making it a robust and equitable benchmark for evaluating multilingual multimodal models.

\section{Evaluation Metrics and the MCQA Framework}
\label{app:mcqaframework}

Traditional evaluation metrics for VLMs, such as exact match accuracy, BLEU~\citep{blue2002}, ROUGE~\citep{lin-2004-rouge}, and CIDEr~\citep{vedantam2015cider}, rely on surface-level n-gram comparisons that often penalize semantically equivalent answers phrased differently from reference texts. In contrast, the multiple-choice question answering (MCQA) framework~\citep{hendrycks2021measuring, romero2024cvqa, LuNEURIPS2022_11332b6b, yue2023mmmu} offers a more human-like evaluation paradigm by providing predefined answer options. This reduces ambiguity in scoring and facilitates the creation of evaluation datasets that capture both domain knowledge and linguistic/cultural nuances across languages. Although concerns regarding oversaturation and reliance on superficial cues in MCQA exist~\citep{du10.1145/3596490, Yuksekgonul2022WhenAW}, these can be mitigated by extending the answer option space and applying rigorous filtering strategies~\citep{WangNEURIPS2024_ad236edc, yue2023mmmu}. Our primary challenge lies in the scarcity of questions that are both multimodal and culturally agnostic. As demonstrated by results from~\benchmark and related studies~\citep{Maaz2024PALOAP, nayak-etal-2024-benchmarking}, oversaturation is not a prevalent issue. Consequently, bridging this evaluation gap is of key importance. To ensure high data quality, source data in \benchmark are manually verified by qualified processors in accordance with established criteria (\ref{quality-assesment}), maintaining a clear distinction between verified and unverified data.

\subsection{Selected Dataset Samples}
\label{sec:selected-data-samples}
The subsequent table presents one sample from each dataset, including the question, the associated image, the provided answers, and the correct answer highlighted in green.



\section*{Supplementary material}

| Language | Question Image | Question and Answer |
|---|---|---|
| Arabic | 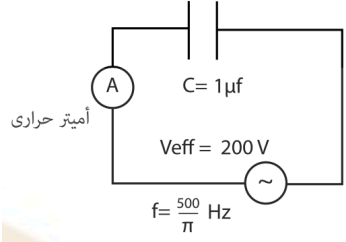 | High School Exam — Physics<br><br>**Question:** قراءة الأميتر الحراري في الدائرة الموضحة تساوي ......<br>**Options:**<br>**A. 0.2 A**<br>B. 2 A<br>C. 0.02 A<br>D. 20 A |
| Arabic | 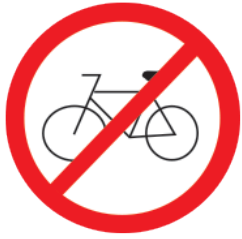 | High School Exam — Geology<br><br>**Question:** من خلال الجدول استنتج الدولة الأقل تضرراً حال تعرضها لزلزال<br>**Options:**<br>A. ل<br>**B. ع**<br>C. ص<br>D. س |
| Bengali | 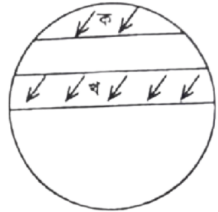 | BRTA Driving Test — Driving<br><br>**Question:** এই চিহ্নটি দ্বারা কি বুঝায়?<br>**Options:**<br>A. শুধুমাত্র সাইকেল চলাচলের জন্য<br>**B. সাইকেল চলাচল নিষেধ**<br>C. মোটরসাইকেল চলাচল নিষেধ<br>D. শুধুমাত্র মোটরসাইকেল চলাচলের জন্য |
| Bengali | | HSC Exam — Geography<br><br>**Question:** উদ্দীপকের 'ক' ও 'খ' বায়ুপ্রবাহের সাধারণ বৈশিষ্ট্য- i. দক্ষিণ-পশ্চিম দিকে প্রবাহিত হয় ii. ডান দিকে বেঁকে যায় iii. সম উষ্ণতাবিশিষ্ট<br>নিচের কোনটি সঠিক?<br>**Options:**<br>A. i ও ii<br>B. i ও iii<br>**C. ii ও iii**<br>D. i, ii ও iii |
| Bengali | 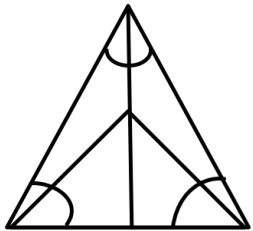 | BCS Exam — Reasoning<br><br>**Question:** নিচের চিত্রে কয়টি ত্রিভুজ আছে?<br>**Options:**<br>A. ৫টি<br>B. ৬টি<br>**C. ৮টি**<br>D. ৪টি |

| Language | Question Image | Question and Answer |
|---|---|---|
| Dutch | 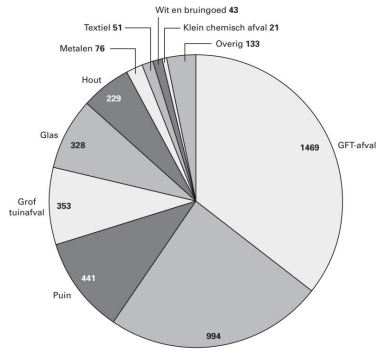 | Dutch Central Exam — Economics<br><br>**Question:** Bekijk bovenstaand diagram. Hoeveel ton klein chemisch afval werd er in Nederland in 2000 ingezameld?<br>**Options:**<br>**A. "21.000 ton"**<br>B. "19.500 ton"<br>C. "23.000 ton"<br>D. "20.500 ton" |
| English | 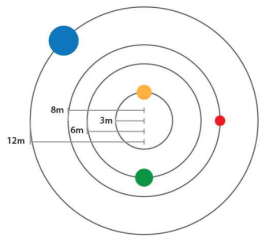 | SAT — Mathematics<br><br>**Question:** In the equation, p and t are constants. Which of the following could be the value of p?<br>**Options:**<br>**A. 2**<br>B. 3<br>C. 4<br>D. 9 |
| English | 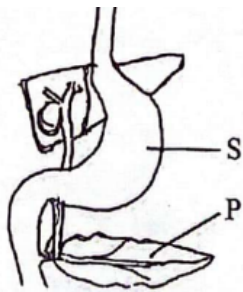 | UCEED Exam — Design<br><br>**Question:** Four spheres start revolving clockwise in concentric circles from their initial positions as shown below. Yellow travels at 2m/sec, green at 4m/sec, red at 2m/sec and blue at 4m/sec. Which of the following statement(s) is/are TRUE?<br>**Options:**<br>**A. Yellow and green never cross (overtake) each other**<br>B. Red and blue takes the same time to complete one revolution<br>C. Yellow takes less time than green to complete one revolution<br>D. Blue and red will cross each other twice after the first 3 complete revolutions of blue |
| English |  | HSC Exam — Biology<br><br>**Question:** Which type of food digest in lebelled 'S' mentioned in the figure?<br>**Options:**<br>A. Potato<br>**B. Pulse**<br>C. Oil<br>D. Ghee |

| Language | Question Image | Question and Answer |
|---|---|---|
| English | 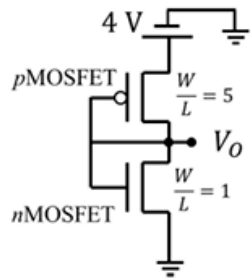 | GATE — Engineering <br><br>**Question:** Consider the CMOS circuit shown in the figure (substrates are connected to their respective sources). The gate width $W$ to gate length $L$ ratios $W/L$ of the transistors are as shown. Both transistors have the same gate oxide capacitance per unit area. For the pMOSFET, the threshold voltage is $-1V$ and the mobility of holes is $40\,\text{cm}^2/\text{V.s}$. For the nMOSFET, the threshold voltage is $1V$ and the mobility of electrons is $300\,\text{cm}^2/\text{V.s}$. The steady-state output voltage $V_o$ is _____ <br>**Options:** <br>A. equal to 0 V <br>B. more than 2 V <br>**C. less than 2 V** <br>D. equal to 2V |
| Flemish | 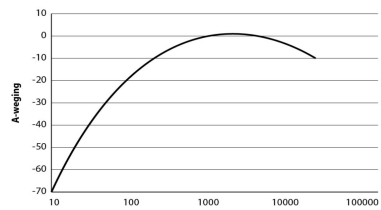 | Physician Exam — Language <br><br>**Question:** Figuur 1A toont de A-weging voor het menselijk gehoor. Deze figuur leert ons dat <br>**Options:** <br>**A. de mens tonen rond de 1000 Hz het beste hoort.** <br>B. mensen tonen van 10.000 Hz niet meer kunnen horen. <br>C. een toon met dezelfde fysische geluidssterkte altijd even intens wordt gehoord. <br>D. bij gelijke geluidssterkte, een mens 100 Hz zachter hoort dan 50 Hz. |
| French | 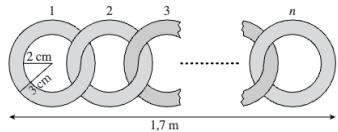 | Mathematical Kangaroo — Mathematics <br><br>**Question:** On attache ensemble des anneaux comme indiqué ci-contre de façon à former une chaîne de 1,7 m de longueur. Combien d'anneaux sont nécessaires ? <br>**Options:** <br>A. 30 <br>**B. 42** <br>C. 21 <br>D. 85 |
| German | 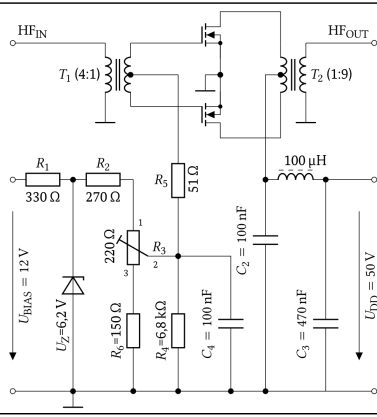 | Amateur Radio Exam — Engineering <br><br>**Question:** Wie groß ist die Gate-Source-Spannung, wenn sich der Schleifer von $R_3$ am Anschlag 1 befindet? <br>**Options:** <br>**A. 3,5 V** <br>B. 2,77 V <br>C. 3,7 V <br>D. 0,45 V |

| Language | Question Image | Question and Answer |
|---|---|---|
| Hindi | 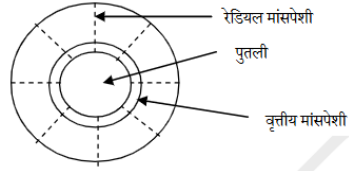 | Science Olympiad  Biology<br><br>**Question:** अजय अपने घर से 20 मिनट पैदल चलकर दोपहर के 3.30 बजे सिनेमा हॉल पहुँचा। वह जल्दी-जल्दी सिनेमा हॉल में घुस गया। इसे अस-पास साफ-साफ देखने में कुछ समय लगा। इस दौरान उसकी आँखों में किस तरह के परिवर्तन आए होंगे?<br><br>**Options:**<br>A. वृत्तीय और रेडियल मांसपेशियां शिथिल होती हैं जबकि पुतलियां संकुचित होती हैं।<br>**B. वृत्तीय मांसपेशियां शिथिल होती हैं, रेडियल मांसपेशियां संकुचित होती हैं और पुतलियां फैलती हैं।**<br>C. वृत्तीय और रेडियल मांसपेशियां संकुचित होती हैं जबकि पुतलियां फैलती हैं।<br>D. वृत्तीय मांसपेशियां संकुचित होती हैं, रेडियल मांसपेशियां शिथिल होती हैं और पुतलियां संकुचित होती हैं। |
| Hindi | 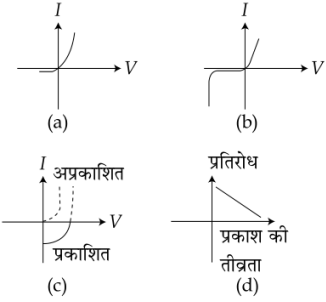 | JEE (Main)  Physics<br><br>**Question:** चित्र (a), (b), (c), (d) देखकर निर्धारित करें कि ये चित्र क्रमशः किन सेमीकंडक्टर डिवाइसों के अभिलक्षणांक ग्राफ हैं :<br>**Options:**<br>**A. साधारण डायोड, जीनर डायोड, सोलर सेल, LDR (लाइट डिपेंडेंट रेजिस्टेंस)**<br>B. जीनर डायोड, साधारण डायोड, LDR (लाइट डिपेंडेंट रेजिस्टेंस), सोलर सेल<br>C. सोलर सेल, LDR (लाइट डिपेंडेंट रेजिस्टेंस), जीनर डायोड, साधारण डायोड<br>D. जीनर डायोड, सोलर सेल, साधारण डायोड, LDR (लाइट डिपेंडेंट रेजिस्टेंस) |
| Hindi | 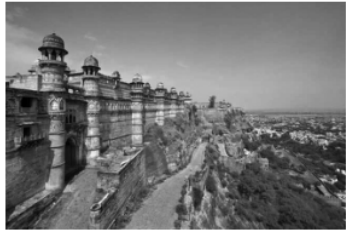 | UP-CET  Social Sciences<br><br>**Question:** चित्र में दिए किले को पहचानिये।<br>**Options:**<br>A. जोधपुर किला<br>**B. ग्वालियर किला**<br>C. लाल किला<br>D. आमेर किला |
| Hindi | $\int_1^2 (x^2 - 2x + 4)^{3/2}\, dx = \dfrac{k}{k+5}$ | JEE (Main)  Mathematics<br><br>**Question:** यदि इस छवि में दिखाए गए समीकरण के अनुसार, तो $k$ बराबर है :<br>**Options:**<br>A. 1<br>B. 2<br>**C. 3**<br>D. 4 |

| Language | Question Image | Question and Answer |
|---|---|---|
| Hindi | 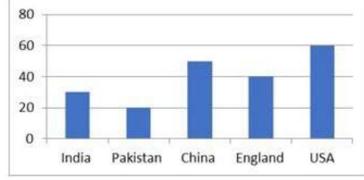 | SSC CGL Exam — Reasoning<br><br>**Question:** विभिन्न देशों की औद्योगिक वृद्धि (₹ करोड़ में) में निम्नलिखित में से कितने देशों की औद्योगिक वृद्धि औसत औद्योगिक वृद्धि से अधिक है?<br>**Options:**<br>A. 4<br>**B. 2**<br>C. 3<br>D. 1 |
| Lithuanian | 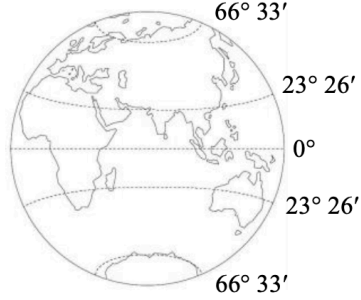 | High School Exam — Geography<br><br>**Question:** Kiek platumos laipsnių yra tarp Šiaurės poliarinio rato ir Pietų atogrąžos?<br>**Options:**<br>A. Apie 23°.<br>**B. Apie 90°.**<br>C. Apie 100°.<br>D. Apie 132°. |
| Nepali | 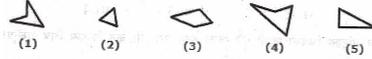 | PSC Exam — Reasoning<br><br>**Question:** दिइएको चित्र १,२,३,४ र ५ मध्येबाट कुनै तीन चित्रहरु एक आपसमा मिलाउदा पुर्ण आकारको त्रिभुजको चित्र बन्दछ । उक्त पूर्ण आकारको बनाउने चित्रहरुको नम्बरहरु दिइएको विकल्पबाट छनौट गर्नुहोस् ।<br>**Options:**<br>A. 124<br>**B. 234**<br>C. 245<br>D. 345 |
| Persian | 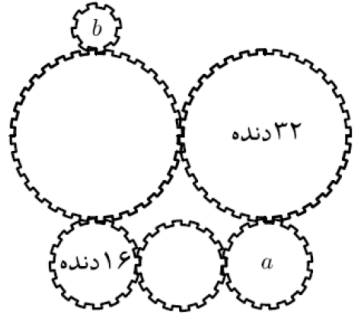 | Olympiad of Informatics — Math<br><br>**Question:**<br>طبق شکل زیر تعدادی چرخ‌دنده داریم که با هم درگیر هستند. چند دور و در کدام جهت باید چرخ دندهی b را بچرخانیم تا چرخ‌دندهی a دقیقا یک دور ساعت گرد بچرخد؟ تعداد دنده‌های چرخ‌دنده‌ی کوچک ۸، چرخ‌دنده‌های متوسط ۱۶ و چرخ‌دنده‌های بزرگ ۳۲ است.<br>**Options:**<br><br>A. 1 دور ساعت گرد<br>B. 1 دور پادساعت گرد<br>C. 2 دور ساعت گرد<br>**D. نمی‌توان چرخ‌دنده‌ی a را چرخاند** |

| Language | Question Image | Question and Answer |
|---|---|---|
| Persian | 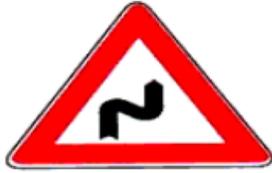 | Driving Test / Driving<br>**Question:** عنوان این تابلو چیست؟<br>**Options:**<br>A. پیچ های پی در پی (اولین پیچ به چپ)<br>**B. پیچ های پی در پی (اولین پیچ به راست)**<br>C. پیچ به راست<br>D. پیچ به چپ |
| Persian | 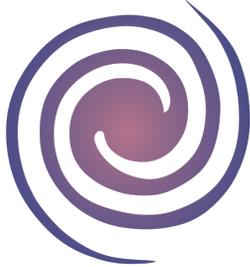 | High School Exam / Engineering<br>**Question:** شکل مقابل نمایانگر کدام است؟<br>**Options:**<br>A. قانون دوم کپلر<br>B. نظریهٔ مه‌بانگ<br>**C. کهکشان راه شیری**<br>D. راه مکه |
| Persian | 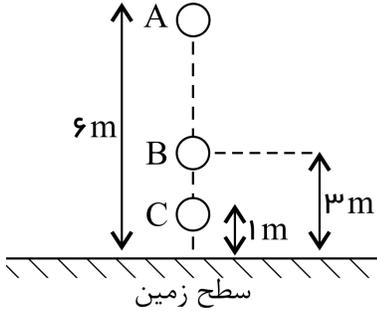 | University Entrance Exam / Physics<br>**Question:** گلوله‌ای مسیری مطابق شکل را طی می‌کند. کار نیروی وزن از A تا B چند برابر کار نیروی وزن از B تا C است؟<br>**Options:**<br>A. 1.5<br>B. 1.25<br>C. 1.2<br>D. 2 |
| Persian | 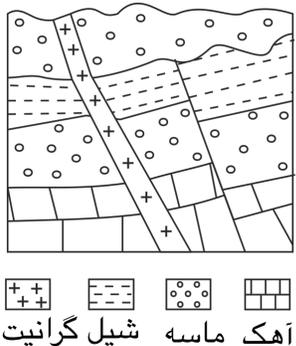 | University Entrance Exam / Geology<br>**Question:** با توجه به گزینه‌ها، در شکل روبرو به ترتیب قدیمی‌ترین و جوان‌ترین پدیده کدام است؟<br>**Options:**<br>A. پیشروی دریا - گسل عادی<br>**B. پسروی دریا - تزریق دایک**<br>C. پیشروی دریا - ناپیوستگی هم‌شیب<br>D. پسروی - گسل عادی |

| Language | Question Image | Question and Answer |
|---|---|---|
| Portuguese | 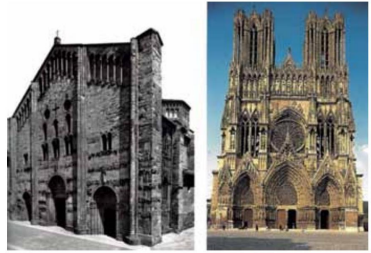 | UNESP — Social Sciences<br>**Question:** Observe as fachadas de duas igrejas. À esquerda, a Basílica de San Michele, construída no século XII em Pavia, na Itália. À direita, a Catedral de Reims, erguida a partir do século XIII em Reims, na França.<br>(Georges Duby e Michel Laclotte (orgs.). História artística da Europa: a Idade Média II, 1998.)<br>As duas fachadas<br>**Options:**<br>**A. diferenciam-se pela pouca ornamentação de San Michele, que expressa o estilo românico, e pela monumentalidade e sofisticação de Reims.**<br>B. diferenciam-se pela solidez de San Michele, que simboliza a força espiritual do catolicismo, e pela carência de detalhes na sede papal em Reims.<br>C. igualam-se na suntuosidade e no rebuscamento arquitetônico, indicando o poderio econômico da Igreja católica.<br>D. diferenciam-se pela discrição de San Michele, que revela o rigor na conduta dos protestantes, e pela ostentação da riqueza católica de Reims. |
| Portuguese | 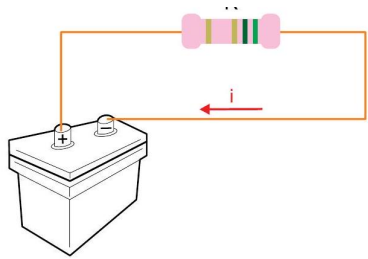 | FAMERP Entrance Exam — Physics<br>**Question:** Quando um gerador de força eletromotriz 12 V é ligado a um resistor R de resistência $5,8\,\Omega$, uma corrente elétrica i de intensidade 2,0 A circula pelo circuito.<br>R<br>A resistência interna desse gerador é igual a<br>**Options:**<br>A. $0,40\,\Omega$.<br>**B. $0,20\,\Omega$**<br>C. $0,10\,\Omega$.<br>D. $0,30\,\Omega$. |
| Portuguese | 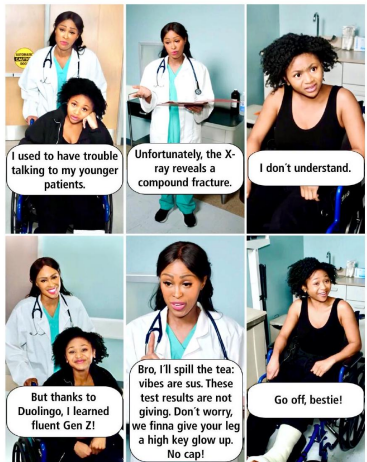 | Unicamp Entrance Exam — Language<br>**Question:** A imagem a seguir apresenta a transcrição de um diálogo em um vídeo publicado no Instagram.<br>No diálogo, a principal característica da reformulação da fala da médica é a inserção de<br>**Options:**<br>A. expressões que utilizam verbos frasais para recontextualizar o tratamento da paciente.<br>B. abreviações de substantivos, através das quais a médica amplia as informações do caso.<br>C. gírias que utilizam diversas classes de palavras para especificar melhor o diagnóstico da paciente.<br>**D. vocábulos marcados pela oralidade, através dos quais a médica atualiza os procedimentos futuros.** |

| Language | Question Image | Question and Answer |
|---|---|---|
| Portuguese |  | [ENEM, Brazil] [Mathematics] <br>**Question:** Um segmento de reta está dividido em duas partes na proporção áurea quando o todo está para uma das partes na mesma razão em que essa parte está para a outra. Essa constante de proporcionalidade é comumente representada pela letra grega $\varphi$, e seu valor é dado pela solução positiva da equação $\varphi^2 = \varphi + 1$.<br>Assim como a potência $\varphi^2$, as potências superiores de $\varphi$ podem ser expressas da forma $a\varphi + b$, em que $a$ e $b$ são inteiros positivos, como apresentado no quadro.<br>A potência $\varphi^7$, escrita na forma $a\varphi + b$ ( $a$ e $b$ são inteiros positivos), é<br>**Options:**<br>A. $7\varphi + 2$<br>B. $9\varphi + 6$<br>C. $11\varphi + 7$<br>**D. $13\varphi + 8$** |
| Serbian | 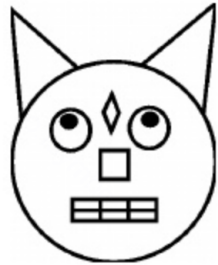 | [Mathematical Kangaroo] [Mathematics] <br>**Question:** Колико процената површине троугла на слици је осенчено?<br>**Options:**<br>**A. 88 %**<br>B. 90 %<br>C. 85 %<br>D. 80 % |
| Russian | 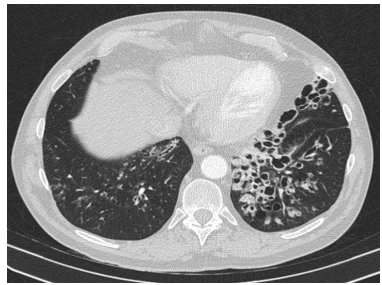 | [Mathematical Kangaroo] [Mathematics] <br>**Question:** Каких геометрических фигур нет на рисунке?<br>**Options:**<br>A. кругов<br>B. все эти фигуры есть<br>C. прямоугольников<br>**D. треугольников** |
| Spanish | | [Medicine Exam] [Pulmonology] <br>**Question:** Varón de 60 años, fumador activo, que presenta tos y expectoración diaria de años de evolución, ocasionalmente hemoptoica. En los últimos meses se añade disnea progresiva. Presenta acropaquia y en la auscultación pulmonar destacan roncus y sibilantes teleinspiratorios en pulmón izquierdo. La TC pulmonar de alta resolución se muestra en la imagen adjunta. ¿Cuál es el diagnóstico más probable?<br><br>**Choices:**<br>A. Carcinoma quístico.<br>B. Enfisema pulmonar.<br>C. Tuberculosis cavitada.<br>**D. Bronquiectasias.** |

| Language | Question Image | Question and Answer |
|---|---|---|
| Spanish | | [Undergraduate Exam] [Biophysics] <br><br> **Question:** Calcule el valor de la primera resistencia (R1) <br> **Options:** <br> A. 42 Ω <br> B. 6 Ω <br> **C. 12 Ω** <br> D. 24 Ω |
| Spanish | | [High School Exam, Colombia] [Biology] <br><br> **Question:** En un laboratorio se estudia el comportamiento del volumen de un gas ideal al variar su temperatura, obteniendo la siguiente gráfica: Teniendo en cuenta la información de la gráfica, si la temperatura aumenta de -153 °C a -33 °C, ¿qué pasa con el volumen del gas? <br> **Options:** <br> A. Disminuye de 30 L a 25 L. <br> B. Disminuye de 10 L a 5 L. <br> C. Aumenta de 0 L a 10 L. <br> **D. Aumenta de 10 L a 20 L.** |
| Telugu | | [Undergraduate Exam] [Chemistry] <br><br> **Question:** ఇచ్చిన చిత్రంలో సమ్మేళనం యొక్క మోలార్ ద్రవ్యరాశి ఎంత? <br> **Options:** <br> A. 304.9 <br> B. 304.4 <br> C. 301.9 <br> **D. 303.4** |
| Ukrainian | | [ZNO Vision] [Mathematics] <br><br> **Question:** Пластикові кульки радіуса 6 см зберігають у висувній шухлядці, що має форму прямокутного паралелепіпеда (див. рисунок). Якою з наведених може бути висота $h$ цієї шухлядки? <br> **Options:** <br> A. 3 см <br> B. 6 см <br> C. 10 см <br> **D. 13 см** |
| Ukrainian | | [Driving Test] [Driving] <br><br> **Question:** По якій траєкторії можна продовжити рух праворуч легковому автомобілю? <br> **Options:** <br> A. Тільки по А. <br> **B. Тільки по Б.** <br> C. По А і Б. <br> D. По будь-якій. |

Table 16: Samples from various exams in the KALEIDOSCOPE benchmark. The correct answer is highlighted in **Bold Green**. Some samples are reformatted for better presentation.



\end{document}

%% file: tables/complete_lang_acc.tex
\begin{table}[H]
\resizebox{\textwidth}{!}{%
\begin{tabular}{@{}clllllllllllllllllll@{}}
\toprule
\multicolumn{1}{l}{} &
   &
  \multicolumn{6}{c}{\textbf{Latin Script}} &
  \multicolumn{12}{c}{\textbf{Non-Latin Script}} \\ 
  \cmidrule(l){3-8} \cmidrule(l){9-20}
\multicolumn{1}{l}{} &
   &
  English &
  French &
  German &
  Dutch &
  Portuguese &
  Spanish &
  Arabic &
  Bengali &
  Croatian &
  Hindi &
  Hungarian &
  Lithuanian &
  Nepali &
  Persian &
  Russian &
  Serbian &
  Telugu &
  Ukrainian \\ \midrule
\multirow{3}{*}{\textbf{Gemini 1.5 Pro}} &
  Total Acc. &
  62.7 &
  54.6 &
  52.6 &
  61.5 &
  81.8 &
  78.5 &
  44.5 &
  51.8 &
  46.9 &
  62.6 &
  39.1 &
  75.0 &
  22.2 &
  41.2 &
  45.0 &
  41.9 &
  58.1 &
  70.3 \\
 &
  Valid Acc. &
  63.2 &
  55.2 &
  52.6 &
  62.5 &
  83.4 &
  78.5 &
  44.7 &
  52.7 &
  48.1 &
  63.2 &
  40.5 &
  75.0 &
  22.8 &
  42.1 &
  46.2 &
  43.6 &
  58.3 &
  70.3 \\
 &
  FE &
  0.9 &
  1.0 &
  0.0 &
  1.6 &
  1.9 &
  0.0 &
  0.5 &
  1.8 &
  2.5 &
  0.9 &
  3.4 &
  0.0 &
  2.4 &
  2.1 &
  2.6 &
  3.8 &
  0.4 &
  0.0 \\ \midrule
\multirow{3}{*}{\textbf{Claude 3.5 Sonnet}} &
  Total Acc. &
  36.9 &
  51.7 &
  73.7 &
  65.2 &
  83.8 &
  77.6 &
  50.3 &
  49.5 &
  46.9 &
  57.1 &
  38.8 &
  81.2 &
  27.8 &
  45.6 &
  45.3 &
  38.9 &
  56.0 &
  75.2 \\
 &
  Valid Acc. &
  63.2 &
  51.7 &
  73.7 &
  65.6 &
  83.8 &
  77.7 &
  50.3 &
  49.6 &
  47.5 &
  57.2 &
  38.9 &
  81.4 &
  28.0 &
  45.6 &
  45.4 &
  39.6 &
  56.0 &
  75.2 \\
 &
  FE &
  41.6 &
  0.0 &
  0.0 &
  0.6 &
  0.0 &
  0.1 &
  0.0 &
  0.2 &
  1.2 &
  0.1 &
  0.4 &
  0.3 &
  0.8 &
  0.0 &
  0.2 &
  1.8 &
  0.0 &
  0.0 \\ 
  \midrule
\multirow{3}{*}{\textbf{GPT-4o}} &
  Total Acc. &
  42.6 &
  46.2 &
  52.4 &
  60.1 &
  76.6 &
  73.8 &
  41.9 &
  60.2 &
  36.4 &
  53.0 &
  36.4 &
  76.2 &
  19.0 &
  41.3 &
  37.6 &
  32.4 &
  41.6 &
  68.5 \\
 &
  Valid Acc. &
  64.5 &
  49.2 &
  53.7 &
  66.8 &
  81.0 &
  78.6 &
  49.7 &
  62.8 &
  40.4 &
  59.0 &
  40.7 &
  79.7 &
  20.5 &
  42.6 &
  40.3 &
  38.7 &
  47.9 &
  77.5 \\
 &
  FE &
  33.9 &
  6.0 &
  2.5 &
  10.0 &
  5.4 &
  6.1 &
  15.7 &
  4.0 &
  9.9 &
  10.1 &
  10.5 &
  4.4 &
  7.1 &
  3.0 &
  6.7 &
  16.2 &
  13.2 &
  11.6 \\ \midrule
\multirow{3}{*}{\textbf{Qwen2.5-VL-72B}} &
  Total Acc. &
  53.8 &
  46.2 &
  49.3 &
  57.4 &
  73.3 &
  72.5 &
  36.1 &
  49.5 &
  33.3 &
  46.2 &
  37.5 &
  69.1 &
  23.8 &
  35.4 &
  39.3 &
  35.9 &
  47.0 &
  65.2 \\
 &
  Valid Acc. &
  53.8 &
  46.4 &
  49.3 &
  57.5 &
  73.3 &
  72.5 &
  36.1 &
  49.5 &
  33.3 &
  46.2 &
  37.6 &
  69.1 &
  23.8 &
  35.4 &
  39.3 &
  35.9 &
  47.0 &
  65.2 \\
 &
  FE &
  0.0 &
  0.5 &
  0.0 &
  0.2 &
  0.0 &
  0.0 &
  0.0 &
  0.0 &
  0.0 &
  0.0 &
  0.2 &
  0.0 &
  0.0 &
  0.0 &
  0.0 &
  0.0 &
  0.0 &
  0.0 \\\midrule
\multirow{3}{*}{\textbf{Aya-Vision-32B}} &
  Total Acc. &
  32.9 &
  27.6 &
  39.1 &
  47.5 &
  57.4 &
  53.2 &
  30.4 &
  26.0 &
  29.6 &
  32.9 &
  26.1 &
  48.5 &
  22.2 &
  30.3 &
  29.0 &
  28.4 &
  34.8 &
  47.1 \\
 &
  Valid Acc. &
  33.0 &
  29.3 &
  39.5 &
  48.7 &
  58.8 &
  55.5 &
  30.5 &
  26.1 &
  29.6 &
  33.7 &
  26.5 &
  48.5 &
  22.2 &
  31.1 &
  29.4 &
  28.5 &
  34.8 &
  47.1 \\
 &
  FE &
  0.4 &
  6.0 &
  1.1 &
  2.4 &
  2.4 &
  4.2 &
  0.5 &
  0.2 &
  0.0 &
  2.3 &
  1.6 &
  0.0 &
  0.0 &
  2.5 &
  1.3 &
  0.4 &
  0.1 &
  0.0 \\ \midrule
\multirow{3}{*}{\textbf{Aya-Vision-8B}} &
  Total Acc. &
  28.9 &
  30.2 &
  32.4 &
  42.2 &
  47.6 &
  46.3 &
  27.2 &
  18.2 &
  29.0 &
  29.2 &
  27.1 &
  34.4 &
  23.8 &
  27.7 &
  25.2 &
  28.5 &
  11.1 &
  36.6 \\
 &
  Valid Acc. &
  28.9 &
  30.2 &
  32.4 &
  42.2 &
  47.6 &
  46.3 &
  27.2 &
  25.9 &
  29.0 &
  29.3 &
  27.1 &
  34.4 &
  24.2 &
  27.8 &
  25.2 &
  28.5 &
  25.7 &
  36.8 \\
 &
  FE &
  0.0 &
  0.0 &
  0.0 &
  0.0 &
  0.0 &
  0.0 &
  0.0 &
  29.5 &
  0.0 &
  0.2 &
  0.0 &
  0.0 &
  1.6 &
  0.3 &
  0.0 &
  0.0 &
  56.8 &
  0.6 \\  \midrule
\multirow{3}{*}{\textbf{Molmo-7B-D}} &
  Total Acc. &
  26.9 &
  33.1 &
  19.9 &
  41.5 &
  47.6 &
  47.8 &
  21.5 &
  26.0 &
  30.9 &
  30.0 &
  25.5 &
  35.3 &
  25.4 &
  28.3 &
  29.2 &
  27.2 &
  33.9 &
  35.8 \\
 &
  Valid Acc. &
  27.0 &
  33.3 &
  19.9 &
  41.5 &
  47.6 &
  47.8 &
  21.5 &
  26.0 &
  30.9 &
  30.1 &
  25.5 &
  35.3 &
  25.4 &
  28.3 &
  29.2 &
  27.2 &
  33.9 &
  35.8 \\
 &
  FE &
  0.2 &
  0.8 &
  0.0 &
  0.0 &
  0.0 &
  0.0 &
  0.0 &
  0.0 &
  0.0 &
  0.2 &
  0.0 &
  0.0 &
  0.0 &
  0.0 &
  0.0 &
  0.0 &
  0.0 &
  0.1 \\ \midrule
\multirow{3}{*}{\textbf{Pangea-7B}} &
  Total Acc. &
  24.7 &
  25.5 &
  17.2 &
  37.3 &
  48.8 &
  46.2 &
  20.4 &
  27.8 &
  17.9 &
  24.3 &
  23.9 &
  32.6 &
  17.5 &
  21.2 &
  25.5 &
  26.4 &
  18.6 &
  33.0 \\
 &
  Valid Acc. &
  27.9 &
  31.1 &
  19.6 &
  39.0 &
  51.0 &
  49.4 &
  22.3 &
  31.8 &
  21.0 &
  29.7 &
  26.2 &
  33.4 &
  23.9 &
  25.2 &
  28.9 &
  30.1 &
  33.9 &
  33.9 \\
 &
  FE &
  11.4 &
  18.1 &
  12.2 &
  4.3 &
  4.3 &
  6.5 &
  8.4 &
  12.8 &
  14.8 &
  18.3 &
  8.8 &
  2.4 &
  27.0 &
  16.0 &
  11.9 &
  12.4 &
  45.2 &
  2.6 \\ \midrule
\multirow{3}{*}{\textbf{Qwen2.5-VL-7B}} &
  Total Acc. &
  36.4 &
  35.2 &
  26.6 &
  46.4 &
  59.2 &
  57.8 &
  36.1 &
  35.0 &
  25.9 &
  33.8 &
  28.6 &
  50.3 &
  22.2 &
  30.3 &
  28.8 &
  27.5 &
  37.6 &
  45.4 \\
 &
  Valid Acc. &
  36.4 &
  35.3 &
  26.6 &
  46.4 &
  59.2 &
  57.8 &
  36.1 &
  35.1 &
  26.1 &
  33.9 &
  28.6 &
  50.3 &
  22.2 &
  30.3 &
  28.8 &
  27.5 &
  37.6 &
  45.4 \\
 &
  FE &
  0.0 &
  0.3 &
  0.0 &
  0.0 &
  0.0 &
  0.0 &
  0.0 &
  0.2 &
  0.6 &
  0.2 &
  0.2 &
  0.0 &
  0.0 &
  0.0 &
  0.0 &
  0.0 &
  0.0 &
  0.0 \\   \midrule
  \multirow{3}{*}{\textbf{Qwen2.5-VL-3B}} &
  Total Acc. &
  35.0 &
  32.3 &
  21.1 &
  41.8 &
  52.2 &
  53.3 &
  33.5 &
  32.8 &
  25.9 &
  31.3 &
  28.2 &
  35.6 &
  23.8 &
  30.3 &
  29.0 &
  27.8 &
  31.8 &
  40.1 \\
 &
  Valid Acc. &
  35.0 &
  32.4 &
  21.1 &
  41.8 &
  52.3 &
  53.3 &
  33.7 &
  32.8 &
  26.4 &
  31.3 &
  29.2 &
  35.6 &
  23.8 &
  30.4 &
  29.0 &
  28.1 &
  31.8 &
  40.1 \\
 &
  FE &
  0.0 &
  0.3 &
  0.0 &
  0.0 &
  0.1 &
  0.0 &
  0.5 &
  0.2 &
  1.9 &
  0.0 &
  3.2 &
  0.0 &
  0.0 &
  0.2 &
  0.0 &
  1.1 &
  0.0 &
  0.0 \\ \bottomrule
\end{tabular}%
}
\caption{\textbf{Total Accuracy \%}, \textbf{Valid Accuracy \%} and \textbf{Format Error \% (FE)} grouped by Language in \benchmark for multimodal samples.}
\label{tab:complete-lang-acc}
\end{table}

%% file: tables/complete_cat_acc.tex
\begin{table}[H]
\resizebox{\textwidth}{!}{%
\begin{tabular}{@{}clllllllllllllll@{}}
\toprule
\multicolumn{1}{l}{} &
   &
    \thead{\textbf{Biology}} & 
    \thead{\textbf{Chemistry}} & 
    \thead{\textbf{Driving} \\ \textbf{License}} & 
    \thead{\textbf{Economics}} & 
    \thead{\textbf{Engineering}} & 
    \thead{\textbf{Geography}} & 
    \thead{\textbf{History}} & 
    \thead{\textbf{Language}} & 
    \thead{\textbf{Mathematics}} & 
    \thead{\textbf{Medicine}} & 
    \thead{\textbf{Physics}} & 
    \thead{\textbf{Reasoning}} & 
    \thead{\textbf{Social} \\ \textbf{Sciences}} & 
    \thead{\textbf{Sociology}}\\ 
    \midrule
\multirow{3}{*}{\textbf{Gemini 1.5 Pro}}    & Total Acc. & 60.1 & 60.2 & 64.4 & 64.1 & 57.0 & 72.8 & 78.7 & 83.5 & 46.9 & 69.6 & 57.4 & 51.2 & 85.7 & 92.3 \\
                                            & Valid Acc. & 60.3 & 60.4 & 64.4 & 64.1 & 57.3 & 72.8 & 78.7 & 83.5 & 48.6 & 70.2 & 57.8 & 52.0 & 85.7 & 92.3 \\
                                            & FE         & 0.3  & 0.4  & 0.0  & 0.0  & 0.4  & 0.0  & 0.0  & 0.0  & 3.5  & 0.8  & 0.7  & 1.7  & 0.0  & 0.0  \\ \midrule
\multirow{3}{*}{\textbf{Claude 3.5 Sonnet}} & Total Acc. & 61.6 & 59.0 & 64.2 & 63.4 & 50.0 & 81.4 & 83.7 & 85.1 & 43.7 & 72.9 & 58.4 & 52.2 & 82.9 & 91.0 \\
                                            & Valid Acc. & 62.9 & 59.7 & 64.2 & 63.8 & 64.4 & 81.5 & 83.7 & 85.5 & 44.4 & 73.8 & 58.7 & 53.3 & 82.9 & 93.4 \\
                                            & FE         & 2.1  & 1.2  & 0.0  & 0.8  & 22.4 & 0.2  & 0.0  & 0.5  & 1.5  & 1.2  & 0.6  & 2.0  & 0.0  & 2.6  \\ \midrule
\multirow{3}{*}{\textbf{GPT-4o}}            & Total Acc. & 60.7 & 47.1 & 65.5 & 64.1 & 45.7 & 76.2 & 70.8 & 78.3 & 39.4 & 64.6 & 51.9 & 43.9 & 74.3 & 87.2 \\
                                            & Valid Acc. & 63.9 & 52.9 & 73.1 & 66.7 & 56.3 & 80.4 & 86.4 & 85.8 & 44.0 & 75.6 & 54.7 & 51.0 & 88.1 & 93.2 \\
                                            & FE         & 5.1  & 11.1 & 10.4 & 3.8  & 18.9 & 5.2  & 18.1 & 8.7  & 10.5 & 14.6 & 5.1  & 13.9 & 15.7 & 6.4  \\ \midrule
\multirow{3}{*}{\textbf{Qwen2.5-VL-72B}}    & Total Acc. & 37.6 & 33.2 & 39.0 & 37.4 & 28.8 & 40.6 & 48.9 & 72.2 & 30.1 & 36.7 & 33.7 & 27.4 & 52.9 & 64.1 \\
                                            & Valid Acc. & 37.6 & 33.2 & 39.0 & 37.7 & 28.9 & 40.7 & 48.9 & 72.2 & 30.4 & 36.7 & 33.7 & 27.4 & 52.9 & 64.1 \\
                                            & FE         & 0.0  & 0.0  & 0.0  & 0.8  & 0.2  & 0.2  & 0.0  & 0.0  & 0.9  & 0.0  & 0.0  & 0.0  & 0.0  & 0.0  \\ \midrule
\multirow{3}{*}{\textbf{Aya-Vision-32B}}    & Total Acc. & 40.2 & 34.6 & 47.1 & 29.8 & 34.6 & 50.5 & 61.4 & 71.0 & 28.8 & 52.1 & 31.8 & 27.5 & 64.3 & 70.5 \\
                                            & Valid Acc. & 40.7 & 34.8 & 47.1 & 29.8 & 34.8 & 50.5 & 61.4 & 71.2 & 29.6 & 52.3 & 33.0 & 27.6 & 64.3 & 70.5 \\
                                            & FE         & 1.3  & 0.5  & 0.0  & 0.0  & 0.7  & 0.0  & 0.0  & 0.2  & 2.8  & 0.4  & 3.6  & 0.3  & 0.0  & 0.0 \\ \midrule
\multirow{3}{*}{\textbf{Aya-Vision-8B}}     & Total Acc. & 53.8 & 50.0 & 54.5 & 58.8 & 48.4 & 70.4 & 77.1 & 84.9 & 40.3 & 63.3 & 42.3 & 42.3 & 80.0 & 87.2 \\
                                            & Valid Acc. & 53.8 & 50.0 & 54.5 & 58.8 & 48.4 & 70.4 & 77.1 & 85.1 & 40.3 & 63.3 & 42.3 & 42.3 & 80.0 & 87.2 \\
                                            & FE         & 0.0  & 0.0  & 0.0  & 0.0  & 0.0  & 0.0  & 0.0  & 0.2  & 0.1  & 0.0  & 0.0  & 0.0  & 0.0  & 0.0  \\ \midrule
\multirow{3}{*}{\textbf{Molmo-7B-D}}        & Total Acc. & 35.2 & 33.5 & 44.9 & 27.5 & 24.4 & 37.6 & 42.1 & 60.1 & 28.7 & 40.4 & 26.7 & 27.5 & 51.4 & 60.3 \\
                                            & Valid Acc. & 35.3 & 33.5 & 44.9 & 27.5 & 24.4 & 37.6 & 42.1 & 60.1 & 28.8 & 40.4 & 26.7 & 27.5 & 52.2 & 61.0 \\
                                            & FE         & 0.1  & 0.0  & 0.0  & 0.0  & 0.1  & 0.0  & 0.0  & 0.0  & 0.1  & 0.0  & 0.0  & 0.0  & 1.4  & 1.3  \\ \midrule
\multirow{3}{*}{\textbf{Pangea-7B}}         & Total Acc. & 30.9 & 22.0 & 38.2 & 32.1 & 21.4 & 35.6 & 42.1 & 65.8 & 24.8 & 35.0 & 24.7 & 21.9 & 50.0 & 55.1 \\
                                            & Valid Acc. & 33.4 & 34.1 & 39.4 & 33.9 & 24.2 & 36.7 & 42.4 & 66.0 & 29.0 & 38.4 & 28.9 & 26.6 & 53.8 & 57.3 \\
                                            & FE         & 7.5  & 35.3 & 2.9  & 5.3  & 11.5 & 3.0  & 0.6  & 0.2  & 14.4 & 8.8  & 14.7 & 17.6 & 7.1  & 3.8  \\ \midrule
\multirow{3}{*}{\textbf{Qwen2.5-VL-7B}}     & Total Acc. & 34.3 & 16.2 & 41.2 & 22.1 & 30.3 & 38.7 & 45.3 & 56.6 & 28.6 & 35.4 & 27.1 & 24.3 & 57.1 & 57.7 \\
                                            & Valid Acc. & 35.4 & 28.0 & 41.6 & 30.9 & 30.3 & 39.5 & 45.6 & 56.6 & 28.6 & 35.4 & 27.1 & 25.1 & 58.0 & 57.7 \\
                                            & FE         & 3.0  & 42.2 & 1.1  & 28.2 & 0.0  & 1.9  & 0.6  & 0.0  & 0.3  & 0.0  & 0.0  & 3.4  & 1.4  & 0.0  \\ \midrule
\multirow{3}{*}{\textbf{Qwen2.5-VL-3B}}     & Total Acc. & 42.6 & 38.5 & 44.9 & 42.7 & 32.4 & 51.0 & 52.9 & 75.7 & 30.1 & 45.8 & 34.7 & 29.5 & 68.6 & 73.1 \\
                                            & Valid Acc. & 42.6 & 38.5 & 44.9 & 42.7 & 32.4 & 51.0 & 52.9 & 75.7 & 30.1 & 45.8 & 34.7 & 29.5 & 68.6 & 73.1 \\
                                            & FE         & 0.0  & 0.1  & 0.0  & 0.0  & 0.0  & 0.0  & 0.0  & 0.0  & 0.1  & 0.0  & 0.1  & 0.0  & 0.0  & 0.0  \\ 
\bottomrule
\end{tabular}%
}
\caption{\textbf{Total Accuracy \%}, \textbf{Valid Accuracy \%} and \textbf{Format Error \% (FE)} grouped by Subject in \benchmark for multimodal samples.}
\label{tab:complete-cat-acc}
\end{table}

%% file: tables/complete_captioning.tex
\begin{table}[H]
\centering
\begin{tabular}{@{}lllllll@{}}
\toprule
 & \multicolumn{3}{c}{\textbf{Qwen2.5-VL-7B}}              & \multicolumn{3}{c}{\textbf{Gemini 1.5 Pro}}             \\ \cmidrule(l){2-4}  \cmidrule(l){5-7}
 & \textit{Total Acc.} & \textit{Valid Acc.} & \textit{FR} & \textit{Total Acc.} & \textit{Valid Acc.} & \textit{FR} \\ \midrule
Diagram & 37.8 & 37.9 & 0.3 & 58.9 & 59.6 & 1.2 \\
Figure  & 34.6 & 34.8 & 0.7 & 49.0 & 50.0 & 1.9 \\
Graph   & 45.2 & 45.2 & 0.0 & 67.4 & 68.2 & 1.2 \\
Map     & 46.7 & 46.7 & 0.0 & 70.9 & 70.9 & 0.0 \\
Photo   & 53.9 & 54.1 & 0.5 & 74.2 & 74.3 & 0.2 \\
Formula & 37.0 & 37.3 & 0.8 & 66.7 & 68.7 & 2.9 \\
Table   & 34.5 & 34.6 & 0.2 & 74.7 & 76.1 & 1.8 \\
Text    & 79.8 & 79.8 & 0.0 & 83.7 & 83.7 & 0.0 \\ \bottomrule
\end{tabular}%
\caption{\textbf{Total Accuracy \%}, \textbf{Valid Accuracy \%} and \textbf{Format Error \% (FE)} grouped by Image Type in \benchmark for captioning + OCR experiment.}
\label{tab:complete-captioning}
\end{table}